%% file: main.tex
\documentclass[10pt]{article} 
\usepackage[preprint]{tmlr}
\input{math_commands.tex}

\usepackage{hyperref}
\usepackage{url}

\usepackage{times}
\usepackage{latexsym}
\usepackage{tabularx}
\usepackage{booktabs}
\usepackage{multirow}
\usepackage{comment}
\usepackage{amsmath}
\usepackage{graphicx}
\usepackage{enumitem}
\usepackage{tikz}
\usetikzlibrary{arrows.meta,backgrounds}

\definecolor{arcnavy}{HTML}{17365D}
\definecolor{arcblue}{HTML}{E7F0F8}
\definecolor{arcteal}{HTML}{E4F3EF}
\definecolor{arcorange}{HTML}{FCEBD8}
\definecolor{arcpurple}{HTML}{EEE8F5}
\definecolor{arcgray}{HTML}{536273}

\title{A Systematic Study of Small Language Models on Abstract Reasoning Tasks}

\author{\name Nur A Zarin Nishat \email nur.nishat@tib.eu \\
      \addr Leibniz University of Hanover \\
      \addr TIB – Leibniz Information Centre for Science and Technology and University Library 
      \AND
      \name Jens Lehmann \email jlehmnn@amazon.com \\
      \thanks{Work done outside of Amazon.}
      \addr Amazon
      \AND
      \name Andrei Aioanei \email Andrei.Aioanei@tib.eu\\
      \addr TIB – Leibniz Information Centre for Science and Technology and University Library\\
     \AND
      \name Sahar Vahdati\email sahar.vahdati@tib.eu\\
      \addr Leibniz University of Hanover \\
     \addr TIB – Leibniz Information Centre for Science and Technology and University Library}

\begin{document}

\maketitle
\begin{abstract}
Endpoint accuracy on abstract-reasoning benchmarks does not reveal whether a language model has acquired a transferable rule or fit distribution-specific regularities. We study this distinction in small language models on the ARC-TGI benchmark, which organizes abstract grid transformations into controllable task families and supports resampling, spatial shifts, and cross-benchmark transfer. Across more than 1,000 runs, we profile decoder-only, encoder--decoder, and mixture-of-experts model families under supervised fine-tuning. We examine the efficiency and stability of skill acquisition, robustness beyond the training distribution, interactions with model family and task formulation, and layer-wise attention signatures that accompany behavioral differences. Substantial in-distribution accuracy is attainable, but acquisition is sensitive to optimization and unevenly distributed across task families. Performance deteriorates sharply outside the training distribution, including when the rule is retained but grid scale changes. Greater training-set depth and breadth yield uneven gains, while the effect of additional in-context examples depends on model family. Executable-rule induction also yields correct solutions not observed under direct grid generation. On selected tasks, attention diagnostics show distinct concentration and context-dependence profiles, but do not establish general causal mechanisms. Overall, abstract-reasoning scores are conditional on the model, adaptation regime, evaluation distribution, and response format.

\end{abstract}

\section{Introduction}
Language models can produce correct answers to difficult tasks, yet endpoint accuracy alone does not establish whether their success reflects transferable rule acquisition. Performance may instead depend on surface regularities, task-specific training, prompt composition, or the required output representation. Controlled perturbations are therefore needed to distinguish acquisition of a transformation from success under one evaluation condition. This question is especially tractable in small language models (SLMs), for which broad experimental sweeps are feasible while the models remain capable of learning nontrivial transformations.
The Abstraction and Reasoning Corpus (ARC) \cite{chollet2019measure} provides a precise setting for this analysis. Each task presents a few input--output grids that disclose a latent transformation and asks the solver to apply it to a new grid. Exact-match evaluation is unambiguous, and success requires generalization from sparse demonstrations; human studies nevertheless show that the tasks are broadly solvable \cite{legris2024h}. However, the original corpus contains unique tasks rather than controlled distributions of related instances. Concept-organized variants improve diagnosis across spatial and semantic abstractions, but remain fixed collections of puzzles \cite{moskvichev2023conceptarc}. The release of new ARC versions as systems specialize on earlier tasks further underscores the importance of preserving evaluation novelty \cite{chollet2025arc}. The original corpus is consequently well suited to benchmarking but less suited to separating within-family fitting from transfer under parameter, spatial, or novel-family shifts.

We address this limitation with a controlled study using the ARC-TGI benchmark \cite{lehmann2026arctgi}.
This benchmark represents a transformation as a task-family generator that can sample variations in color, position, object configuration, and grid size. 
It supports a hierarchy of evaluation conditions: held-out seen-family episodes resample from training generators; within-family shifts retain the transformation while varying controlled properties; and novel-family or cross-benchmark transfer introduces unseen generators or original ARC evaluation tasks. The experiments reported here instantiate this hierarchy through held-out seen-family episodes, controlled grid-scale shifts, and ARC-AGI cross-benchmark transfer. Generator-disjoint novel-family evaluation remains a capability of the broader protocol but is not analyzed as a separate quantitative endpoint. This hierarchy turns ARC from a single endpoint into an instrument for examining skill acquisition and transfer.
We use this instrument to profile three transformer model families---decoder-only (D), encoder--decoder (ED), and mixture-of-experts (MoE)---across more than 1,000 controlled training and evaluation runs. The study is descriptive rather than a causal comparison of transformer architectures: each family is represented by a different pretrained model, and the observed differences therefore combine architectural, pretraining, tokenization, and capacity effects. Our analysis is organized around the following questions:
\begin{enumerate}[noitemsep]
    \item \textbf{RQ1:} How efficiently and stably do small language models acquire abstract transformation skills?
    \item \textbf{RQ2:} How robustly do acquired skills transfer beyond the training distribution?
    \item \textbf{RQ3:} How do model family and task formulation modulate the capabilities that are expressed?
    \item \textbf{RQ4:} What attention signatures appear in selected diagnostic cases under spatial shift?
\end{enumerate}

This study makes three contributions. First, it defines a controlled ARC-TGI evaluation protocol that separates acquisition, training experience, distribution shift, and task formulation while profiling differences among model families. Second, it applies this protocol in a large behavioral analysis of optimization sensitivity, learning dynamics, training- and inference-time experience, and transfer. The results show that strong seen-family performance can coexist with narrow task-family coverage and weak robustness to distribution shift. Third, attention-entropy analysis and attention-edge intervention provide complementary selected-task diagnostics, which we interpret as exploratory evidence rather than causal explanations of architectural differences. Together, these contributions characterize when abstract-transformation performance is acquired, retained, and expressed under alternative evaluation conditions. 

\section{Related Work}
\textbf{ARC-AGI evaluation and solver design.}
ARC-AGI tests whether a solver can infer novel visual transformations from limited demonstrations \cite{chollet2019measure}. Human studies show that the tasks are challenging but broadly solvable, whereas GPT-family performance varies substantially across task types \cite{legris2024h, mitchell2023comparing}. Later ARC-AGI versions maintain novelty as systems specialize on earlier benchmarks \cite{chollet2025arc, foundation2026arc}; broader evaluations and surveys likewise caution against interpreting leaderboard scores without the task distribution and inference procedure \cite{gendron2024llms, vahdati2026arc}. Consequently, comparisons across ARC systems should identify both task novelty and the amount of solver-side adaptation. Performance also depends on solver design. Object-oriented representations can expose spatial structure hidden by plain-text grids \cite{xu2023llms}, while test-time training, augmentation, and search add task-specific computation \cite{akyurek2024surprising, franzen2024architect, chollet2024arcprize}. These methods allocate computation at different stages---input construction, per-task optimization, or inference-time search---so equal endpoint accuracy need not reflect equal transfer. Correct outputs may coexist with inconsistent explanations \cite{lee2025reasoning}, and direct-grid transduction and executable-rule induction solve overlapping but distinct task subsets \cite{li2025combining}. We therefore treat representation, inference budget, and output formulation as experimental variables rather than interchangeable routes to one score.

\textbf{ARC benchmark and dataset variants.}
ARC-derived resources vary in task organization, response format, and generation
mechanism. ConceptARC groups puzzles by spatial and semantic concepts; Mini-ARC
uses simplified $5\!\times\!5$ grids and supports human interaction traces, and
MC-LARC converts open-ended grid generation into multiple-choice selection
\cite{moskvichev2023conceptarc,kim2023playgrounds,shin2024mclarc}. Data-expansion
suites make a different trade-off: AugARC applies rotational variants to
existing tasks, whereas GIFARC synthesizes analogy-grounded ARC-style tasks
from GIFs \cite{bikov2024augarc,sim2025gifarc}. Procedural suites resample
transformation families: ReARC reconstructs generators for the 400 public ARC
training tasks; ARC-GEN targets exhaustive mimetic coverage of those families,
and ARC-TGI adds episode-level constraints and aligned reasoning and code
artifacts \cite{hodel2024procedural,moffitt2025arcgen,lehmann2026arctgi}. These
resources are complementary but do not yield directly comparable scores: concept grouping,
response-format changes, augmentation, synthesis, and procedural resampling
alter different aspects of the task distribution. Our study requires repeated
family identity and controlled parameters, so it uses ARC-TGI for within-family
interventions and reserves original ARC-AGI tasks for cross-benchmark
evaluation.

\textbf{Controlled generalization and adaptation.}
Controlled perturbations reveal failures hidden by aggregate accuracy. GSM-Symbolic changes surface details while preserving mathematical structure \cite{mirzadeh2025gsm}; complexity-controlled puzzles expose sharp failure regimes whose interpretation can depend on token budgets and output constraints \cite{shojaee2026illusion, opus2025illusion}. Synthetic-data studies separately analyze knowledge, manipulation, reasoning, and planning \cite{AllenZhuLi2023Series}, while grokking shows generalization can emerge after training accuracy saturates \cite{DBLP:journals/corr/abs-2201-02177}. These results motivate reporting trajectories across controlled shifts rather than endpoints alone. A useful evaluation should test whether gains persist as nuisance features, complexity, or spatial scale change while the latent transformation is preserved. Training choices are equally consequential: supervised fine-tuning and reinforcement learning can alter capabilities differently \cite{huan2025does, zhang2025reinforcement}, and LoRA trades a constrained update space for parameter efficiency \cite{DBLP:conf/iclr/HuSWALWWC22}, affecting both acquisition and forgetting \cite{loradan}. Optimizer choice, update rank, and checkpoint selection can therefore change the apparent model orderings.
Our study consequently controls hyperparameter configuration, adaptation method, checkpointing granularity, training data, prompt strategy, and evaluation distribution shift jointly across all three architectures.

\textbf{Model-family and contextual factors.}
Decoder-only models place demonstrations and targets in one autoregressive stream, whereas encoder--decoder models first construct a bidirectional input representation, as in T5 and LongT5 \cite{DBLP:journals/jmlr/RaffelSRLNMZLL20, DBLP:conf/naacl/GuoAUONSY22}. T5Gemma obtains an encoder--decoder model through continued pretraining of decoder models \cite{DBLP:journals/corr/abs-2504-06225}; sparse MoE models instead route tokens through subsets of experts \cite{DBLP:conf/iclr/ShazeerMMDLHD17, DBLP:journals/corr/abs-2101-03961}. Because representatives also differ in pretraining, tokenization, context length, and capacity, we report model-family profiles rather than isolated architectural effects. This distinction is especially important for serialized grids, whose token length grows with spatial scale. Matched spatial and prompting conditions help separate this contextual sensitivity from overall endpoint performance. Because attention weights become difficult to interpret after layer-wise mixing and even attention interventions provide imperfect relevance evidence \cite{abnar-zuidema-2020-quantifying,serrano-smith-2019-attention}, we treat entropy and edge knockout as complementary selected-task diagnostics rather than mechanistic explanations.
\section{Experimental Design}
\begin{figure*}[h!]
  \centering
  \begin{tikzpicture}[
    font=\sffamily,
    stage/.style={draw=black, line width=0.72pt, rounded corners=2pt,
      fill=white, align=center, inner sep=3pt},
    prep/.style={stage, minimum width=2.44cm, minimum height=0.68cm,
      text width=2.15cm, inner xsep=3pt, inner ysep=1.5pt},
    exec/.style={stage, minimum width=1.75cm, minimum height=0.88cm,
      text width=1.58cm},
    flow/.style={-{Stealth[length=1.65mm,width=1.20mm]}, draw=black,
      line width=0.68pt, line cap=round, line join=round},
    guide/.style={draw=black, line width=0.56pt,
      line cap=round, line join=round},
    bypass/.style={-{Stealth[length=1.55mm,width=1.10mm]}, draw=black,
      line width=0.52pt, densely dashed, line cap=round, line join=round}
  ]
    
    \node[stage, minimum width=2.20cm, minimum height=3.35cm,
      text width=1.82cm] (environment) at (0,0) {%
      \scriptsize\bfseries STUDY ENVIRONMENT\\[-1pt]
      \rule{1.62cm}{0.42pt}\\[4pt]
      \normalfont\footnotesize ARC-TGI\\curriculum builder\\[3pt]
      \scriptsize training and evaluation curricula\\[3pt]
      \fontsize{6}{6.8}\selectfont seen $\cdot$ shifted\\novel / cross-benchmark};

    \draw[rounded corners=3pt, line width=0.56pt]
      (1.62,-1.72) rectangle (4.38,1.72);
    \node[prep] (complexity) at (3.00,1.15) {%
      \fontsize{6}{6.8}\selectfont\bfseries COMPLEXITY\\[-1pt]
      \normalfont\fontsize{6}{6.8}\selectfont attributes $\cdot$ difficulty};
    \node[prep] (tasks) at (3.00,0) {%
      \fontsize{6}{6.8}\selectfont\bfseries TASK SETUP\\[-1pt]
      \normalfont\fontsize{6}{6.8}\selectfont sampled $\cdot$ held-out};
    \node[prep] (prompts) at (3.00,-1.15) {%
      \fontsize{6}{6.8}\selectfont\bfseries PROMPT BUILDER\\[-1pt]
      \normalfont\fontsize{6}{6.8}\selectfont transduction $\cdot$ induction};
    \node[anchor=south, inner sep=0pt] at (3.00,1.84)
      {\scriptsize\bfseries PARALLEL PREPARATION};

    \draw[guide] (environment.east) -- (1.45,0);
    \draw[guide] (1.45,-1.15) -- (1.45,1.15);
    \draw[flow] (1.45,1.15) -- (complexity.west);
    \draw[flow] (1.45,0) -- (tasks.west);
    \draw[flow] (1.45,-1.15) -- (prompts.west);
    \fill (1.45,0) circle (0.65pt);

    \draw[guide] (complexity.east) -- (4.55,1.15);
    \draw[guide] (tasks.east) -- (4.55,0);
    \draw[guide] (prompts.east) -- (4.55,-1.15);
    \draw[guide] (4.55,-1.15) -- (4.55,1.15);
    \node[stage, minimum width=1.72cm, minimum height=1.12cm,
      text width=1.50cm] (scheduler) at (5.65,0) {%
      \fontsize{6}{6.8}\selectfont\bfseries CURRICULUM\\SCHEDULER\\[2pt]
      \normalfont\fontsize{6}{6.8}\selectfont matched runs};
    \draw[flow] (4.55,0) -- (scheduler.west);
    \fill (4.55,0) circle (0.65pt);

    \draw[rounded corners=3pt, line width=0.56pt]
      (6.58,-1.68) rectangle (12.82,1.68);
    \node[exec] (factory) at (7.82,0.75) {%
      \scriptsize\bfseries MODEL FACTORY\\[-1pt]
      \normalfont\fontsize{6}{6.8}\selectfont decoder-only\\[-1pt]
      encoder--decoder $\cdot$ MoE};
    \node[exec] (adaptation) at (7.82,-0.75) {%
      \fontsize{6}{6.8}\selectfont\bfseries ADAPTATION\\[-1pt]
      \normalfont\fontsize{6}{6.8}\selectfont full fine-tuning\\[-1pt]
      LoRA};
    \node[stage, minimum width=1.55cm, minimum height=1.18cm,
      text width=1.27cm] (run) at (9.85,0) {%
      \scriptsize\bfseries TRAIN / INFER\\[2pt]
      \normalfont\fontsize{6}{6.8}\selectfont distributed\\matched runs};
    \node[stage, minimum width=1.65cm, minimum height=1.18cm,
      text width=1.35cm] (evaluate) at (11.82,0) {%
      \scriptsize\bfseries EVALUATE\\[2pt]
      \normalfont\fontsize{6}{6.8}\selectfont exact grid match\\generator profiles};
    \node[anchor=south, inner sep=0pt] at (9.70,1.80)
      {\scriptsize\bfseries DISTRIBUTED TRAINING / INFERENCE};

    \draw[guide] (scheduler.east) -- (6.68,0);
    \draw[guide] (6.68,-0.75) -- (6.68,0.75);
    \draw[flow] (6.68,0.75) -- (factory.west);
    \draw[flow] (6.68,-0.75) -- (adaptation.west);
    \fill (6.68,0) circle (0.65pt);
    \draw[guide] (factory.east) -- (8.88,0.75);
    \draw[guide] (adaptation.east) -- (8.88,-0.75);
    \draw[guide] (8.88,-0.75) -- (8.88,0.75);
    \draw[flow] (8.88,0) -- (run.west);
    \fill (8.88,0) circle (0.65pt);
    \draw[flow] (run.east) -- (evaluate.west);

    \node[stage, minimum width=2.22cm, minimum height=3.70cm,
      text width=1.88cm] (analysis) at (14.15,0) {%
      \scriptsize\bfseries RESULT ANALYSIS\\[-1pt]
      \rule{1.68cm}{0.42pt}\\[2pt]
      \scriptsize \textbf{RQ1} Acquisition\\[-1pt]
      \normalfont\fontsize{6}{6.8}\selectfont
      efficiency $\cdot$ stability\\[2pt]
      \scriptsize\textbf{RQ2} Transfer\\[-1pt]
      \normalfont\fontsize{6}{6.8}\selectfont
      robustness under shift\\[2pt]
      \scriptsize\textbf{RQ3} Modulation\\[-1pt]
      \normalfont\fontsize{6}{6.8}\selectfont
      model $\cdot$ experience\\[-1pt]formulation\\[2pt]
      \scriptsize\textbf{RQ4} Case studies\\[-1pt]
      \normalfont\fontsize{6}{6.8}\selectfont
      entropy $\cdot$ edge knockout};
    \draw[flow] (evaluate.east) -- (analysis.west);

    \begin{scope}[on background layer]
      \draw[bypass] (environment.south) -- (0,-1.93) --
        node[fill=white, inner sep=1pt, below=1pt]
        {\fontsize{6}{6.8}\selectfont curriculum specification}
        (5.65,-1.93) -- (scheduler.south);
    \end{scope}
  \end{tikzpicture}%
  \caption{\textbf{Controlled study workflow.} The curriculum builder creates parallel complexity, task, and prompt views that converge in a matched scheduler. The scheduler instantiates model-family and adaptation choices, launches distributed training or inference, and routes behavioral and selected-task diagnostic evidence to the four research questions. The dashed path denotes the curriculum constraints applied directly to scheduling.}
  \label{fig:study-framework}
\end{figure*}
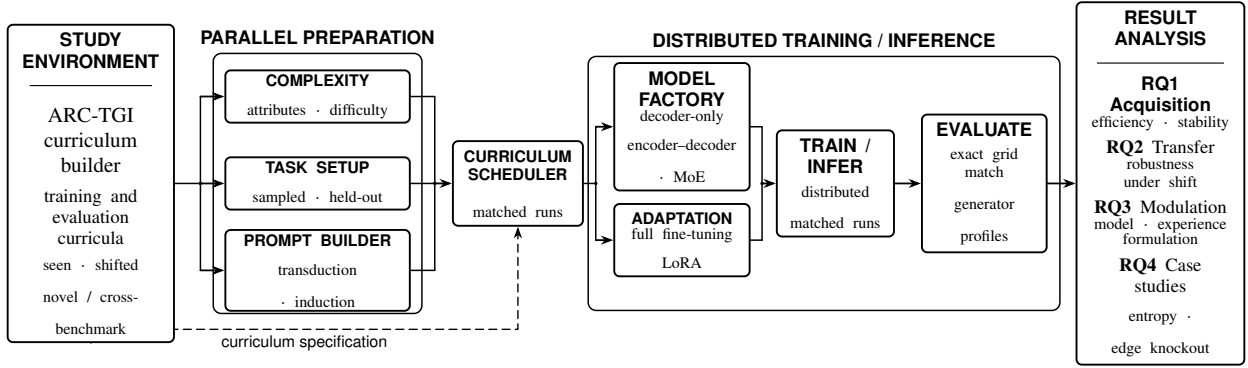

Figure \ref{fig:study-framework} summarizes how the research questions are translated into matched experiments. Each curriculum is analyzed for experiment-specific complexity and converted into task and prompt variants before a common scheduler launches the selected model and adaptation configurations. Distributed training and inference then produce the behavioral results for acquisition, transfer, and formulation, followed by the selected-task internal diagnostics.
Each intervention changes a designated axis---optimization, training experience,
evaluation distribution, demonstration count, or output formulation---while
holding the remaining protocol fixed wherever the comparison permits. We
interpret differences within these matched blocks and reserve cross-family
comparisons for descriptive profiling because the pretrained representatives
are not fully matched. The following subsections describe the controlled
distributions, model and formulation choices, and reporting protocol in this
order.

\subsection{Controlled Task Distributions}
Each ARC task contains several input--output demonstration grids and a test input whose output is determined by the same latent transformation. ARC-AGI provides unique tasks and exact-match evaluation, but it does not provide repeated samples from a common transformation family. ARC-TGI supplies this missing experimental control: a generator deterministically implements a transformation while sampling colors, positions, object configurations, lengths, and grid sizes. Operationally, each generator is a Python module: \texttt{create\_input} samples
a valid grid, \texttt{transform\_input} applies the fixed rule, and
\texttt{create\_grids} assembles demonstrations and a test pair while rejecting
invalid or ambiguous episodes. Task-level variables remain fixed within an
episode, whereas grid-level variables may vary among demonstrations. This enables
resampling surface realizations while retaining the rule; because validity depends
on the implementation, claims are scoped to the curated families.

The released inventory contains 461 human-validated generators covering 180
ARC-Mini, 215 ARC-AGI-1 (200 training and 15 evaluation), and 66
ARC-AGI-2 (55 training and 11 evaluation) tasks \cite{lehmann2026arctgi}. Each generated episode includes grids, instantiated input, transformation reasoning chains, and partially evaluated Python programs.
The same latent task can therefore support direct grid transduction,
executable-rule induction, and reasoning-description prompts without changing
the underlying transformation. Our curricula use only the ARC-AGI-1
training-family subset that satisfies the study-specific constraints;
the exact support for each intervention is
reported below.

\paragraph{Rationale for benchmark selection.}
Benchmark selection was guided by four requirements. First, the evaluation must
distinguish repeated exposure to an individual task from learning within a
transformation family; task-disjoint resamples from a shared generator provide
this comparison. Second, a distribution shift must be able to preserve the rule
while changing a specified observable property; separate task- and grid-level
variables support targeted spatial and surface shifts. The present experiments
instantiate this control through grid scale, while other parameter-level shifts
remain extensions of the protocol. Third, demonstrations
must jointly specify the intended transformation without relying on attributes
that appear only at test time. Episode constraints, executable checks of stored
outputs, optional shortcut screening, and human refinement provide safeguards
for this requirement. Finally, correspondence with source ARC tasks permits a
separate cross-benchmark evaluation on the original ARC-AGI set. These controls
support the required comparisons, but they do not imply that all generators are
equally natural or difficult. We therefore retain generator-level reporting and
scope the conclusions to the filtered families.

\begin{figure}[t]
  \centering
  \includegraphics[width=0.70\linewidth]{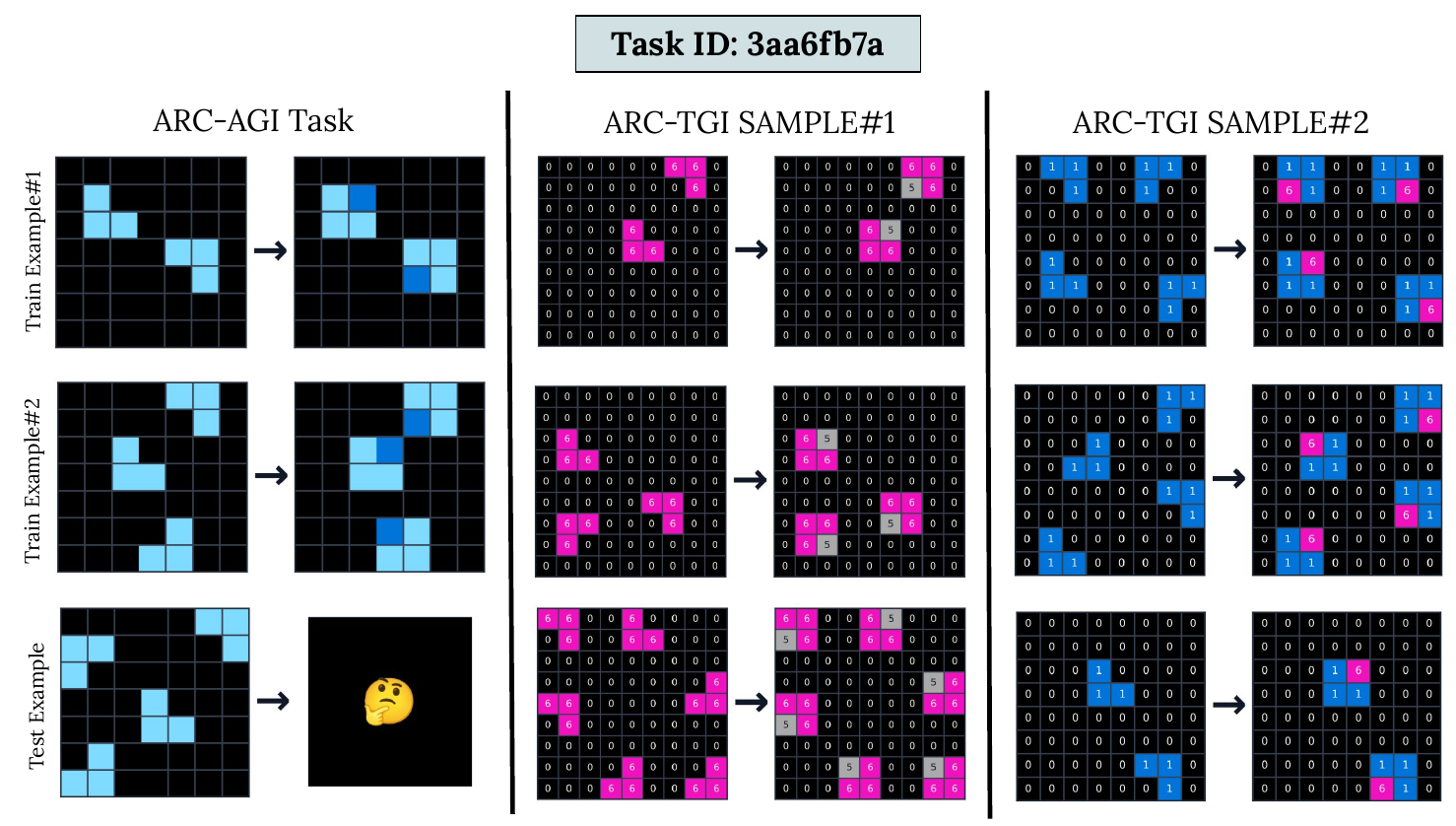}
  \caption{\textbf{Representative ARC-TGI transformation family.} Grid-level variables change object configuration, positions, and colors. Task-level variables determine their joining direction and alignment within each generated episode.}
  \label{fig:arc-tgi-example}
\end{figure}

Figure \ref{fig:arc-tgi-example} illustrates a representative family. The rule
identifies separated colored horizontal segments and joins them into an
aligned vertical stack. Segment count, length, and color vary across grids,
whereas joining direction and horizontal alignment remain fixed across the
demonstrations and test pair of one episode; these task-level choices can change
between generated samples. The samples are therefore not duplicates of the
source puzzle: they preserve the transformation family while requiring the
episode-specific realization to be inferred from its demonstrations. 

\textbf{Held-out seen-family evaluation} resamples new episodes from generators used during training. \textbf{Within-family shifts} retain the transformation but change controlled properties, including generator parameters or grid scale. \textbf{Novel-family evaluation and cross-benchmark transfer} introduce unseen generators or original ARC-AGI tasks. Dataset construction enforces disjoint task identifiers and, where required, disjoint generator sets. Appendix \ref{app:data} describes generation, filtering, and leakage controls. Curricula are filtered by token length, grid size, and experiment-specific condition. Each
record stores the demonstration grids, test grid, task identifier, and generator
identifier. Natural-language grid descriptions and transformation traces are
included only when required by the prompt condition. Training and evaluation
always use disjoint task identifiers, including when they share a generator.
Novel-family evaluation additionally uses generator-disjoint splits, whereas
spatial transfer reuses generators but constrains grid sizes on each side. The
reported results use spatial and cross-benchmark transfer; the generator-disjoint
split is documented as protocol support rather than as a separate empirical
endpoint. Since
the original ARC-AGI evaluation set changes both generator source and
difficulty, it is a cross-benchmark condition rather than a pure intervention
on novelty.

Within-family depth varies the number of training examples per
generator over $n\in\{10,20,40,50,70,80,100\}$ and evaluates every model on a
common disjoint set of 30 examples per generator. Training breadth varies the
number of generators over $\{60,80,100,140,180,193\}$ with 50 training examples
per generator; evaluation uses 30 examples from 50 families shared across all
scales. Spatial transfer uses 123 generators shared across directions with 50
training, and 20 evaluation examples per generator
under matched or crossed
small- and large-grid regimes. Full schedules are reported in Appendices
\ref{app:experience} and \ref{app:generalization}.

\subsection{Models, Training, and Task Formulations}
We study one representative from each of three transformer families. The decoder-only model is Qwen3-1.7B; the encoder--decoder model is T5Gemma-ml-ml-ul2-it with approximately 2B parameters; and the sparse model is Qwen1.5-MoE-A2.7B-Chat, with approximately 14.3B total and 2.7B active parameters. We report comparisons at the \emph{model-family} level because the representatives differ in pretraining, tokenization, context length, and total capacity; the design therefore does not isolate architecture as a causal factor.

Selection required repeated fine-tuning, structured generation, instruction
following, and sufficient context for multiple grids. Qwen3 provides a
32,768-token context, whereas the ED model provides 8K; the MoE
representative matches the intended scale more closely in active parameters than
in total capacity. These differences remain part of the model-family profiles,
not controlled architectural effects. All models are trained with supervised fine-tuning. We evaluate both full-parameter updates and parameter-efficient updates through LoRA \cite{DBLP:conf/iclr/HuSWALWWC22}. The large hyperparameter sweep establishes a suitable operating region for each family before the generalization and experience analyses.
Across 624 configurations, the initial sweep varies learning rate, warm-up ratio,
weight decay, and epoch. The D and MoE sweeps use learning rates
$\{5\!\times\!10^{-6},10^{-5},2\!\times\!10^{-5},2\!\times\!10^{-4}\}$,
whereas the ED sweep uses
$\{10^{-5},5\!\times\!10^{-5},10^{-4},2\!\times\!10^{-4}\}$. All use warm-up
ratios $\{0.03,0.05,0.1\}$; weight decay is $\{0,0.01,0.05,0.1\}$ for the
D and MoE models and $\{0,0.01,10^{-4},10^{-5}\}$ for the
ED model. Checkpoint grids extend to 35 epochs for decoder-only,
30 for ED, and 12 for MoE. LoRA tests
$(r,\alpha)\in\{(16,32),(32,64),(64,128),
(128,256)\}$ with dropout 0.05, keeping $\alpha/r=2$. Configuration selection
is performed within each family; the formulation study additionally repeats a
smaller warm-up and weight-decay search with epochs fixed at 10 for the
D and ED models and five for MoE. The sweep locates
operating regions rather than causal optimizer effects; Appendix
\ref{app:optimization} retains run-level timing,
memory, and cost-adjusted compute tables, whose H200 $=2\times$ H100 conversion
is only an accounting convention.

We consider two output formulations. In \textbf{transduction}, demonstrations and a test input are mapped directly to the output grid. In \textbf{induction}, the model generates a Python function that expresses the inferred transformation and is executed on the test input. Within each formulation, we compare a few-shot prompt with a reasoning-description prompt that supplies natural-language descriptions and transformation traces. This design tests whether formulation changes which capabilities are surfaced rather than assuming that all prompts measure the same behavior.
\begin{figure*}[h!]
  \centering
  \includegraphics[width=\linewidth]{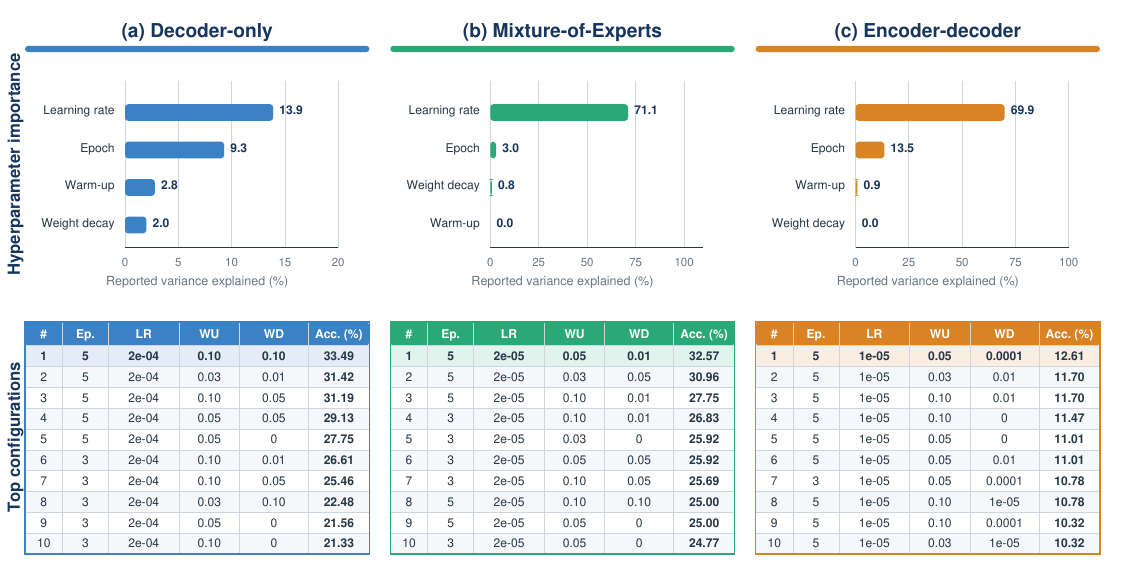}
  \caption{\textbf{RQ1: hyperparameter sensitivity and leading configurations.}
  Top: reported importance scores for the decoder-only, mixture-of-experts, and
  encoder--decoder model families. Bottom: the ten highest-accuracy
  configurations in each sweep. Axes are scaled independently by family, and
  importance scores are descriptive rather than an additive variance
  partition. LR, WU, and WD denote learning rate, warm-up ratio, and weight
  decay.}
  \label{fig:rq1-sensitivity}
\end{figure*}
\begin{figure*}[t!]
  \centering
  \includegraphics[width=\linewidth]{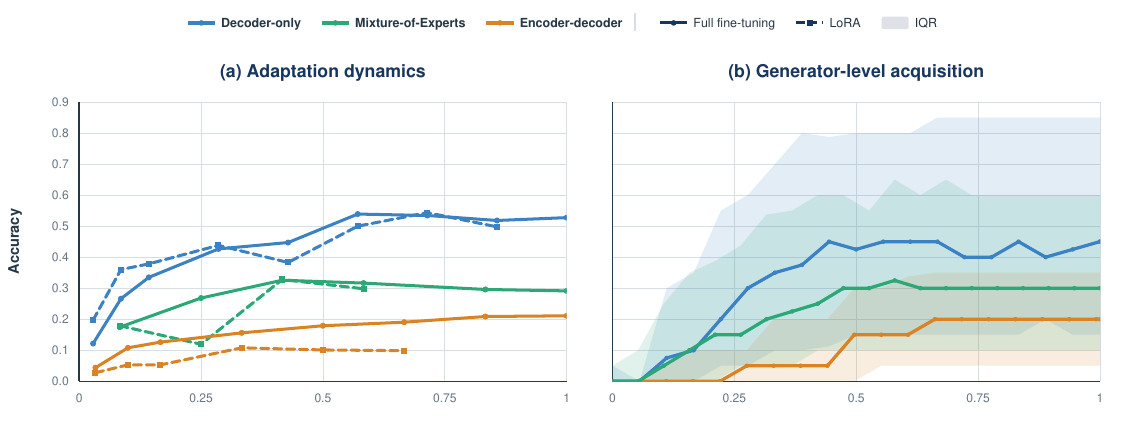}
  \caption{\textbf{RQ1: adaptation and acquisition dynamics.} (a) Exact-match
  accuracy for the selected full fine-tuning and LoRA configurations over
  model-family-specific normalized training progress. Colors identify model
  families; solid circles denote full fine-tuning and dashed squares denote
  LoRA. (b) Median generator-level accuracy with interquartile bands over
  normalized training progress. Endpoint, rank, and compute details appear in
  Appendix \ref{app:optimization}; disaggregated generator trajectories appear
  in Appendix \ref{app:experience}.}
  \label{fig:rq1-dynamics}
\end{figure*}
\subsection{Evaluation and Reporting}
The primary metric is exact grid-match accuracy. We additionally inspect generator-level accuracy to determine whether aggregate improvements extend broadly across families or are driven by a few. Transduction outputs are extracted from structured tags. Induction outputs are parsed as code, executed in a sandboxed environment, and compared with the target grid. Unless explicitly stated, subsequent experiments use the best-performing configurations identified by the tuning study. In that study, every configuration uses the same seed-42 split of the filtered task records: 95\% is used for training and the remaining 5\% for configuration comparison. We therefore treat the 5\% holdout as validation support rather than as an independent final test set. Training does not use a separate evaluation dataset for validation-based checkpoint selection; checkpoint performance is measured after training. Ray coordinates distributed training on the SLURM cluster, while inference uses
actors pinned to individual GPUs. The workloads run on H100 and H200 hardware,
with the allocation recorded for each configuration. 

For MoE, target-module detection distinguishes expert
feed-forward targets, which may require dynamic unused-parameter detection, from
attention-only targets compatible with static-graph execution. The reported MoE
LoRA runs adapt the attention projections. At inference, the decoder-only and
MoE models use vLLM
, whereas the ED model uses the
Hugging Face generation pipeline. The reproducibility protocol requires model
and tokenizer revisions, random seeds,
package versions, launch commands, dataset manifests, and checkpoint-selection
rules. Together, these fields make the data, execution backend, and selected
checkpoint auditable for each reported condition. The study comprises more than 1,000 controlled training and evaluation runs, including 624 hyperparameter configurations. This breadth supports behavioral profiling, but it is not a substitute for uncertainty estimates: the main text therefore emphasizes large, repeated patterns across families and avoids claims unsupported by the design.
RQ4 adds two selected-task diagnostics after behavioral evaluation:
layer-wise attention entropy over the test-input region,
computed for each family over its own attention type, and attention-edge
knockout, which measures the percentage change in mean
per-token target log probability after masking one prompt region's
incoming edges within a five-layer
sliding layer window. Because their effect
scales differ across families, we compare layer-wise signatures rather than
aggregate their magnitudes into a common score.

\section{Results}
Each subsection answers one research question. We first establish the operating
conditions and stability of acquisition, then test transfer, examine interactions
with task formulation and in-context experience, and finally report exploratory
internal diagnostics. The main text presents the comparisons needed for this
sequence; full response distributions, epoch-wise tables, per-generator scaling
heatmaps, compute accounting, and diagnostic methods appear in the corresponding
appendix sections.

\subsection{Acquisition Efficiency and Stability}
This subsection addresses \textbf{RQ1}. Optimization is a prerequisite for
interpreting all subsequent comparisons. Across 624 configurations, learning
rate and training duration are the dominant modulators, whereas warm-up ratio
and weight decay have smaller effects over the tested ranges. Figure
\ref{fig:rq1-sensitivity} shows that their relative influence is
model-family-specific: learning rate and epoch are both consequential for D,
while learning rate dominates the MoE and ED sweeps. The leading configurations
reinforce this separation---D favors the largest tested learning rate over
longer schedules, MoE peaks in a narrower intermediate-rate region, and ED
favors the smallest tested rate. Consequently, transferring one optimization
recipe across families would confound subsequent comparisons. Appendix
\ref{app:optimization} reports the full marginal response distributions and
search grids.

Parameter-efficient adaptation is also family-dependent. Rank-128 LoRA matches
full fine-tuning for D (54.4\% versus 53.9\%), and rank-16 LoRA reaches a
comparable endpoint for MoE; under the tested configuration, however, ED loses
approximately half of its full fine-tuning accuracy. Figure
\ref{fig:rq1-dynamics}(a) shows that these endpoints arise from different
trajectories. D follows nearly the same path under both update methods; MoE's
selected LoRA run is less stable early in training, and the ED gap opens early
and persists. A single adaptation budget is therefore not a neutral default
across model families. Table \ref{tab:main-summary} and Appendix
\ref{app:optimization} report the selected endpoints and compute details.

Panel (b) shifts the unit of analysis from aggregate accuracy to median
generator-level accuracy. D rises most quickly, reaches about 0.45, and then
fluctuates around 0.40--0.45; MoE improves steadily and stabilizes near 0.30; and
ED improves later and plateaus near 0.20. The broad interquartile bands show that
even these distinct median trajectories conceal substantial variation among
generators. Appendix \ref{app:experience} disaggregates the curves into
checkpoint-level generator trajectories and reports the corresponding improved,
decreased, and zero-performing counts.

Increasing experience improves aggregate performance, but unevenly. The depth
intervention holds 200 transformation families fixed while varying training
examples per family from 10 to 100. Its generator-level profiles retain
persistent easy and difficult bands, and individual families can improve,
plateau, or regress as examples are added (Appendix
\ref{app:experience}, Figure \ref{fig:rq1-sample-scaling}). Additional
within-family variation therefore does not uniformly expand the learnable task
set.

The complementary breadth intervention varies the number of training generators
while evaluating the same 50 families. Table \ref{tab:main-summary} reports the
endpoints: scaling from 60 to 193 generators raises accuracy from 39.3\% to
51.6\% for D, from 39.1\% to 45.0\% for MoE, and from 7.5\% to 13.0\% for ED.
Appendix \ref{app:experience} reports all intermediate scales and shows that
these aggregate gains are driven by only some transformation families.

\noindent\textbf{Answer to RQ1.}
The three model families acquire substantial in-distribution skill, but
acquisition is not uniformly efficient or stable across conditions. It depends
strongly on optimization and adaptation choices, follows
family-specific trajectories, and varies substantially across transformation
families. More experience improves aggregate accuracy without
reliably eliminating this heterogeneity; endpoint comparisons should therefore
be conditioned on both the operating point and the training experience.
\begin{figure*}[t]
  \centering
  \includegraphics[width=\linewidth]{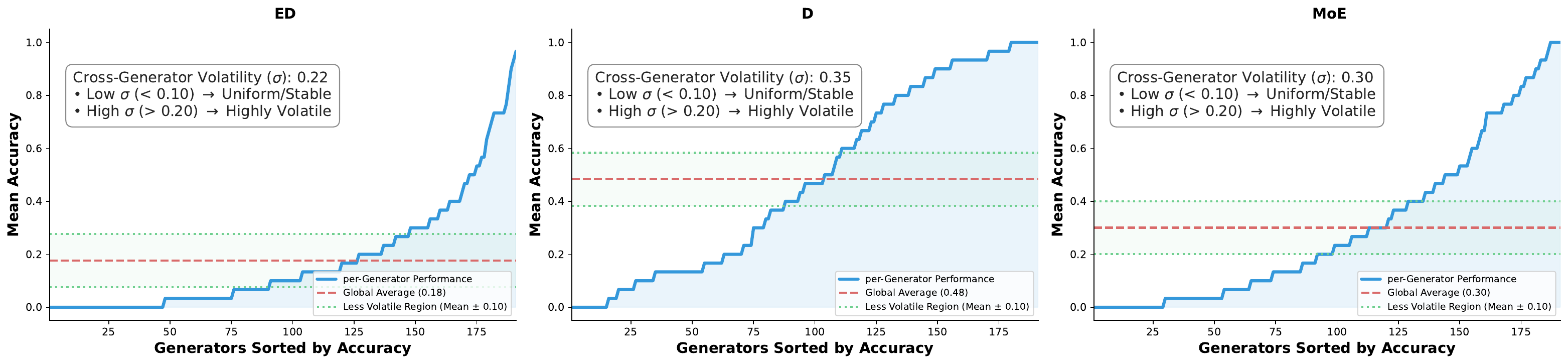}
  \caption{\textbf{RQ2: family-level heterogeneity.} Generators are sorted by exact-match accuracy for each model family. The dashed line is global mean accuracy and dotted lines mark mean $\pm0.1$. Low dispersion can reflect consistent success or uniform failure.}
  \label{fig:rq2-heterogeneity}
\end{figure*}

\begin{table*}[t!]
\centering
\small
\setlength{\tabcolsep}{7pt}
\renewcommand{\arraystretch}{1.02}
\begin{tabular}{@{}lllrrr@{}}
\toprule
\textbf{RQ} & \textbf{Comparison} & \textbf{Condition} & \textbf{D} & \textbf{MoE} & \textbf{ED} \\
\midrule
\multirow{4}{*}{RQ1}
 & \multirow{2}{*}{Training breadth} & 60 generators  & 39.3 & 39.1 & 7.5 \\
 &                                   & 193 generators & 51.6 & 45.0 & 13.0 \\
\cmidrule(lr){2-6}
 & \multirow{2}{*}{Adaptation} & Full fine-tuning & 53.9 & 32.6 & 21.1 \\
 &                              & LoRA             & 54.4 & 32.8 & 10.8 \\
\midrule
\multirow{6}{*}{RQ2}
 & \multirow{2}{*}{Cross-benchmark transfer} & Seen-family val. max.  & 54.4 & 32.8 & 21.1 \\
 &                                            & Observed ARC-AGI peak & 3.5 & 2.75 & 2.00 \\
\cmidrule(lr){2-6}
 & \multirow{2}{*}{Small-grid training} & Small-grid evaluation & 68.5 & 53.1 & 64.4 \\
 &                                      & Large-grid evaluation & 24.3 & 12.7 & 5.9 \\
\cmidrule(lr){2-6}
 & \multirow{2}{*}{Large-grid training} & Large-grid evaluation & 65 & 42.4 & 33.9 \\
 &                                      & Small-grid evaluation & 51.4 & 36.4 & 45 \\
\midrule
\multirow{4}{*}{RQ3}
 & \multirow{2}{*}{Few-shot formulation} & Transduction & 44.3 & 33.7 & 0.0 \\
 &                                        & Induction    & 52.3 & 59.0 & 16.1 \\
\cmidrule(lr){2-6}
 & \multirow{2}{*}{In-context examples} & One shot   & 36.7 & 20.1 & 43.6 \\
 &                                      & Four shots & 51.9 & 30.6 & 23.4 \\
\bottomrule
\end{tabular}
\caption{\textbf{Behavioral results summary.} Exact-match accuracy (\%) under
the two conditions stated for each RQ1--RQ3 comparison. 
Complete trajectories,
intermediate scales, formulation variants, and compute accounting appear in the
corresponding figures and appendices.}
\label{tab:main-summary}
\end{table*}
\subsection{Transfer Beyond the Training Distribution}
This subsection addresses \textbf{RQ2}. The central test is whether improvements on seen task families persist under distribution shift. We evaluate each available training checkpoint on the original ARC-AGI evaluation set. Table \ref{tab:main-summary} (RQ2: Cross-benchmark) contrasts each family's strongest reported seen-family validation performance with its maximum observed ARC-AGI accuracy. Because ARC-AGI is evaluated across multiple checkpoints, these maxima are post-hoc descriptive summaries of the trajectories rather than estimates from checkpoints selected independently of ARC-AGI. The decoder-only model falls from 54.4\% to 3.5\%, MoE from 32.8\% to 2.75\%, and ED from 21.1\% to 2.0\%. Epoch-wise transfer is also non-monotonic even when seen-family accuracy increases (Appendix \ref{app:generalization}). D's performance fluctuates between 0.25\% and 3.5\%; MoE reaches its peak at epochs 1 and 5 but declines between them; ED reaches 2\% at epochs 15 and 20 before falling to 0.25\% at epoch 25. Continued fitting on seen families therefore does not monotonically improve cross-benchmark transfer. Because ARC-AGI differs from ARC-TGI in both novelty and difficulty, this result is best interpreted as cross-benchmark transfer rather than as a pure estimate of generator novelty.

\begin{figure*}[t!]
  \centering
  \includegraphics[width=0.9\linewidth]{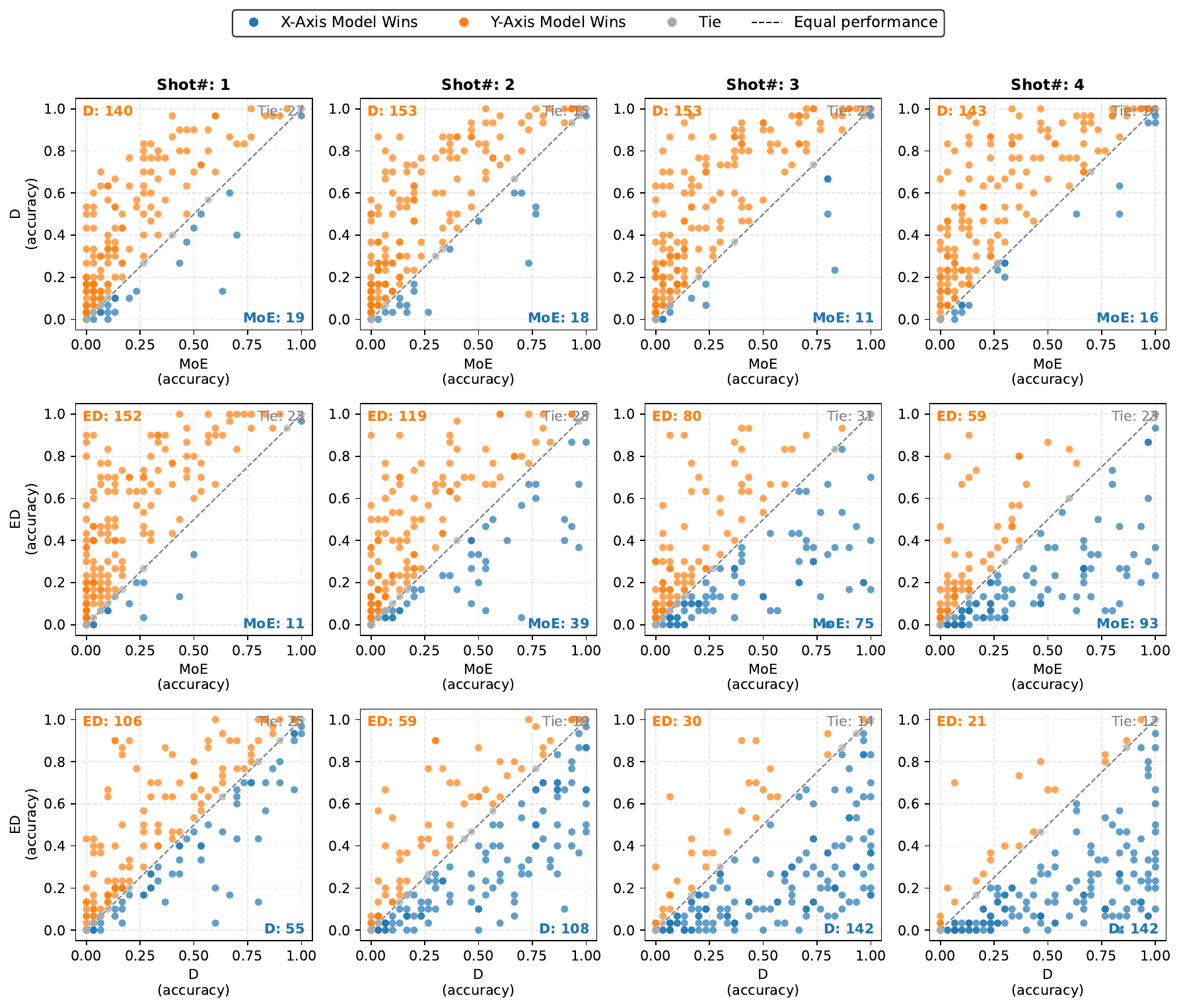}
  \caption{\textbf{RQ3: generator-level effects of in-context experience.} Pairwise generator accuracies for one to four demonstrations. Points above or below the diagonal indicate which model family performs better; gray denotes a tie.}
  \label{fig:rq3-shots}
\end{figure*}
Spatial scale provides a more controlled shift because the transformation family is retained. After filtering, 123 generators are shared across the two directions. Each training curriculum contains 50 samples per generator restricted to small ($\leq15\times15$) or large ($>16\times16$) grids, and each evaluation curriculum contains 20 samples per generator from the opposite or matched scale for 118 out of 123 generators. Table \ref{tab:main-summary} (RQ2: spatial-transfer rows) shows a consistent asymmetry that differs by model. For D and MoE, models trained on small grids perform well on small evaluation grids but collapse in the large-grid regime. Training on larger grids transfers better to small grids, although performance still declines. For D, the matched-to-shifted gap is 44.2 points after small-grid training but 13.6 points after large-grid training. The corresponding gaps are 40.4 versus 6 points for MoE. ED follows this pattern in only one direction: its matched-to-shifted gap after small-grid training is 58.5 points, the largest collapse of the three models. The opposite pattern appears when trained on large grids, where it improves by 11.1 points when transferred to smaller grids. ED therefore shows the strongest asymmetry in both directions. Its shorter context and model-specific input representation are plausible contributors, but the comparison cannot separate those factors from the other model-specific differences. Thus, even when the latent transformation is unchanged, token length and spatial realization materially affect transfer.

The aggregate transfer gaps must also be interpreted relative to the coverage of the seen-family reference score. We sample up to 30 disjoint evaluation instances from each of 191 seen generators and aggregate exact-match accuracy by family. Figure \ref{fig:rq2-heterogeneity} sorts generators by accuracy and reports cross-generator dispersion. High performance is concentrated in roughly 50, 25, and 10 generators for D, MoE, and ED, respectively, while many families remain near zero. D has the highest mean accuracy but also the largest reported cross-generator standard deviation ($\sigma=0.35$); ED has the lowest mean and the smallest dispersion ($\sigma=0.22$), which is driven partly by uniform failure. Dispersion alone is therefore not a sufficient measure of consistency. The profiles show that a high aggregate reference score need not represent broad transformation-family coverage.

\noindent\textbf{Answer to RQ2.}
Acquired performance transfers poorly as evaluation departs from the training
distribution. The degradation appears under cross-benchmark transfer and under
a controlled spatial shift that retains the transformation family; its
directional asymmetry further reveals sensitivity to the surface realization
and length of the serialized task. Seen-family endpoint performance is therefore a
weak proxy for transfer. This evidence is consistent with distribution-specific
fitting, but it does not establish the absence of abstract reasoning because
exact-match behavior alone cannot determine the internal computation.

\subsection{Effects of Model Family and Task Formulation}
This subsection addresses \textbf{RQ3}. No model family dominates every condition. Because the best configuration from direct transduction does not transfer reliably across prompts, the formulation study performs a smaller search over warm-up ratio and weight decay while fixing the epoch budget for each family. With direct transduction, D achieves the strongest accuracy, followed by MoE and ED. Induction changes this ordering: few-shot induction raises D and MoE to 52.3\% and 59\% respectively, substantially narrowing their gap, while ED improves from 0.0\% in few-shot transduction to 16.1\% in few-shot induction. Providing natural-language reasoning information has little or negative benefit for D and MoE under transduction, while it substantially improves ED from 0.0\% to 15.6\%. Under induction, however, reasoning reduces accuracy for D and ED and leaves MoE essentially unchanged. The formulation is therefore not a neutral output wrapper; it changes the computational problem and the prior supplied by the output language.

The number of demonstrations produces another family-specific interaction. Task and generator identifiers are held fixed across $k\in\{1,2,3,4\}$, yielding approximately 5,800 evaluation instances from 186 generators at each comparison. D and MoE improve from one to four shots (36.7\% to 51.9\% and 20.1\% to 30.6\%, respectively), whereas ED declines from 43.6\% to 23.4\%. This inversion is consistent with an input-length or representation bottleneck for the selected encoder--decoder model, but the present design cannot attribute it solely to encoder--decoder architecture.

Table \ref{tab:main-summary} consolidates these formulation and context-length endpoints with the RQ1 and RQ2 comparisons discussed above; the complete formulation and shot-count results appear in Appendix Table \ref{tab:rq3-formulation}.

Aggregate shot accuracy conceals which transformation families drive the reversal. Figure \ref{fig:rq3-shots} compares the families pairwise. At one shot, ED outperforms D on 106 generators and in aggregate accuracy (43.6\% versus 36.7\%). By four shots, D outperforms ED on 142 generators, whereas ED leads on only 21; aggregate accuracy likewise favors D (51.9\% versus 23.4\%). D also increasingly outperforms MoE as demonstrations increase, while MoE outperforms ED in the four-shot condition. The result is not a uniform offset between families: prompt length changes the set of transformations on which each model is competitive.

\noindent\textbf{Answer to RQ3.}
Measured performance depends on the model, training configuration, and task
formulation rather than on the model alone. D has
the strongest direct-transduction performance; MoE often acquires skills
quickly, and ED is competitive in the one-shot setting but degrades with
additional context. Executable-rule induction reveals correct solutions not
observed through direct grid generation, while more demonstrations help the
selected decoder-based models and hurt the selected encoder--decoder model.
These profiles motivate hypotheses about autoregressive processing, sparse
routing, and bidirectional encoding, but they do not support a universal or
causal architecture ranking; that test requires multiple matched models per
family.

\begin{figure*}[!t]
  \centering
  \includegraphics[width=0.86\linewidth]{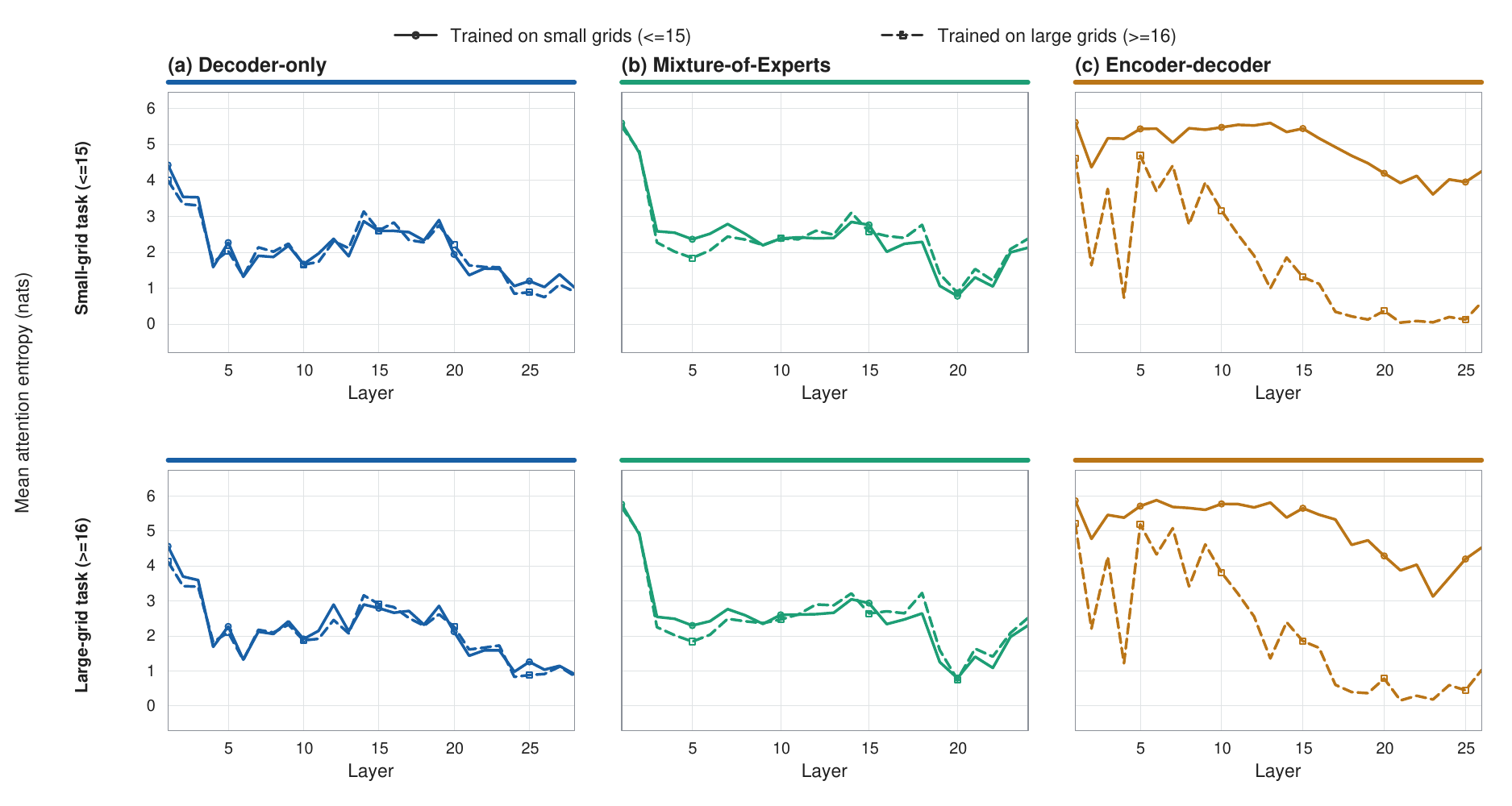}\\[-2pt]
  \includegraphics[width=0.86\linewidth]{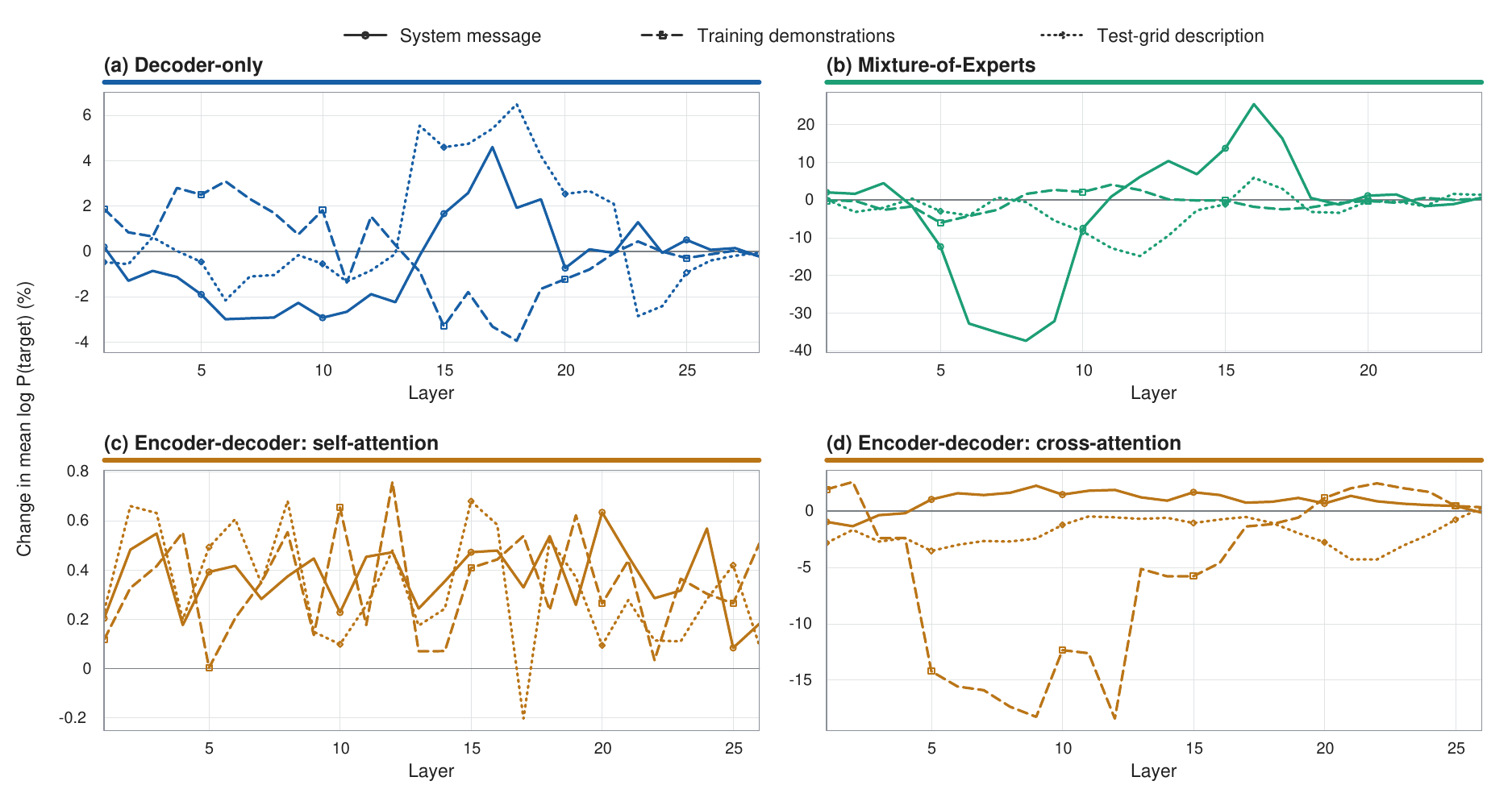}
  \caption{\textbf{RQ4: exploratory layer-wise attention diagnostics.}
  \emph{Top:} mean attention entropy on the selected small-grid task (upper
  row) and large-grid task (lower row) after small-grid or large-grid training.
  \emph{Bottom:} the baseline-normalized change in mean per-token target-answer log probability after blocking attention from
  the system message, demonstrations, or test-grid description. Negative
  knockout values indicate supportive context. Colors identify D (blue), MoE
  (green), and ED (orange); knockout panels retain native scales.}
  \label{fig:rq4-diagnostics}
\end{figure*}

\subsection{Exploratory Internal-Signature Case Studies}
This subsection addresses \textbf{RQ4}. We use two complementary selected-task
diagnostics to examine, in selected cases, 
whether the behavioral differences
under spatial shift are accompanied by distinct layer-wise attention signatures.
The analysis is
deliberately contrastive: for entropy, we select one small-grid and one
large-grid task that models from all three families solve after large-grid
training but fail after small-grid training. The knockout analysis uses a
common selected prompt across families to compare when each prompt region
contributes. The resulting profiles characterize diagnostic cases in which
training scale or model family changes the outcome; they are not
population-level estimates over ARC tasks.

For the entropy diagnostic, we compute Shannon entropy from the normalized
attention distribution, reported as raw per-layer values. The averaging domain differs by family because the diagnostic probes a different attention type in each. For D and MoE, we hook self-attention and average over heads and over query positions in the test-input region. For ED, we hook decoder cross-attention at the first decoder position and average
over heads, with the distribution restricted and renormalized over the encoder key positions in the test-input region. Lower entropy denotes
more concentrated attention, not better reasoning. Figure
\ref{fig:rq4-diagnostics} (top) compares the two selected tasks after training
in either grid regime.  D begins between 4 and 4.4 nats and becomes more
concentrated in its final layers, with the training conditions remaining nearly indistinguishable. MoE starts near 5.5--6 nats, falls rapidly to 2--3 nats,
and exhibits another marked decrease around layer 20. ED remains near
4.6--6.3 nats through much of its cross-attention stack when trained on small-grid tasks, but when trained on large-grid tasks, entropy falls sharply from the middle layers onwards and drops to approximately 1 by the final layers. The two training profiles diverge substantially and are by far the largest separation between the two training regimes of the three families. Entropy thus separates
family-specific concentration profiles, but it cannot identify which prompt
region contributes to the target.

The attention-edge intervention provides a more direct diagnostic. We partition the prompt into system instructions, demonstrations, and the test-grid description. Within each five-layer window, incoming edges from one region are masked (to the final query position for self-attention, and to all decoder query positions for ED cross-attention); the forward pass is rerun without deleting prompt tokens, and the mean per-token target-answer log probability is compared with the unmodified pass. We report this comparison as a percentage change relative to the magnitude of the unmodified value (Appendix~\ref{app:attention-knockout}), so each panel is expressed in units of that model's own baseline confidence. Negative changes indicate that the removed region supported the target; positive changes indicate that it carried a competing signal. Because each panel is normalized to its own baseline, effect magnitudes are not comparable across families; comparisons should focus on sign and layer-wise timing rather than absolute magnitude across families.
The selected D case exhibits a two-phase pattern: demonstrations compete with
the target early but support it after roughly layer 13. The selected MoE case relies strongly on the system message: suppressing it collapses the target probability around layers 5--9 and shows the opposite behavior between layers 10--18, while the model relies on demonstrations in the initial and late layers. ED self-attention is noisier, whereas its decoder cross-attention shows
sustained support from demonstrations through approximately layers 3--17.
Taken together, the diagnostics separate concentration from dependence: ED's
diffuse entropy profile, for example, coexists with a clear cross-attention
dependence on demonstrations. Appendix \ref{app:mechanistic} provides the
complete diagnostic definitions and scope.

\noindent\textbf{Answer to RQ4.}
In the selected diagnostic cases, the model families exhibit distinct internal signatures
in both attention concentration and prompt-region dependence. D and MoE become
more concentrated with depth but rely on demonstrations over different layer
ranges, whereas ED retains diffuse attention and exposes a clearer signal in
decoder cross-attention than in encoder self-attention. These signatures
accompany the behavioral differences, but they neither explain those
differences nor support a causal architectural ranking.

\section{Conclusion}
We used controllable ARC task families to distinguish seen-family acquisition
from performance under shifted evaluation conditions. Across three small-model
families and more than 1,000 controlled runs, in-distribution accuracy could be
substantial, yet acquisition was sensitive to optimization, heterogeneous
across task families, and only weakly predictive of transfer. Increasing
training depth or breadth improved aggregate accuracy unevenly without eliminating
persistent difficult families; spatial and cross-benchmark shifts produced sharp
degradation; and induction, transduction, and demonstration count surfaced
different model-family profiles. Attention diagnostics 
revealed family-specific internal signatures, although they neither explain the
behavioral differences nor support causal architectural rankings.

Taken together, the results support a broader standard of evidence in
abstract-reasoning research. A single score collapses several scientifically
distinct questions: what was acquired, how efficiently and stably it was
acquired, which task families support it, how far it transfers, and which
formulation makes it observable. Resampleable task-family distributions make
these questions experimentally separable. We therefore encourage future ARC
studies to pair endpoint accuracy with learning trajectories, family-level
coverage, matched distribution shifts, and formulation sensitivity, while
using internal diagnostics to generate hypotheses rather than support general
mechanistic claims. Under this view, ARC provides a controlled testbed for
studying the conditions under which measured performance appears, persists, and fails.
Small models are especially useful in this setting: their limitations make
broad controlled interventions feasible and expose structure that a single
frontier-model result can conceal. The goal is not merely to solve more
instances drawn from a familiar distribution, but to acquire transformations
quickly, preserve them across nuisance and spatial changes, and express them
reliably under alternative interfaces. This study does not establish the
absence of abstract reasoning; it shows how to demand more precise evidence for
it. Progress toward general abstract reasoning should therefore be assessed not
only by success rate, but also by acquisition efficiency, task-family coverage,
stability under controlled shifts, and evidence that an acquired transformation
survives those shifts.

\section*{Limitations}
The study has five primary limitations. First, one pretrained model represents each architectural family; pretraining data, tokenizer, context length, parameter count, alignment, and inference backend are therefore confounded with architecture. Second, the sweep prioritizes broad experimental coverage but does not estimate variability across training seeds. Its 5\% holdout supports configuration selection rather than an independent final test, and the reported ARC-AGI maxima are post-hoc summaries across evaluated checkpoints. Third, cross-benchmark transfer to ARC-AGI changes both task-family novelty and difficulty, so it is not a pure OOD intervention. Fourth, exact grid match measures behavioral success but cannot determine whether the internal computation constitutes reasoning; executable code also supplies a strong formal-language prior. Finally, the mechanistic diagnostics cover selected tasks and yield effects on different scales across model families. They are exploratory and should not be read as general causal explanations.

These limitations restrict the conclusions to the selected small-model families, training regimes, and textual ARC representation. They also motivate matched-family comparisons, repeated-seed estimates for headline results, controlled novel-generator splits, and object-structured representations in future work.

\section*{Broader Impact Statement}
This work studies evaluation methodology and small-model behavior rather than a deployed system. Its positive impact is to encourage more cautious interpretation of reasoning benchmarks and more compute-efficient controlled studies. Potential negative impacts include overgeneralizing results from selected models or treating behavioral failures as definitive claims about intelligence. We mitigate these risks by stating the model-family confounds, separating behavioral evidence from mechanistic interpretation, and reporting the limits of cross-benchmark transfer. The experimental campaign also consumed substantial GPU resources; the compute tables support transparent accounting and motivate reuse of released artifacts where possible.

\bibliography{tmlr}
\bibliographystyle{tmlr}

\appendix

\section{Data Construction and Study Infrastructure}
\label{app:data}
The appendix follows the evidence flow of the main paper. We first document the
data and execution controls, then provide the extended RQ1 acquisition analyses,
the RQ2 transfer analyses, the RQ3 formulation study, and finally the RQ4
exploratory internal-signature diagnostics. Within each block, the experimental
setup precedes the detailed results, followed by the interpretation and limits of
the comparison. This organization preserves the distinction used in the main
text between behavioral outcomes, transfer under distribution shift, and
selected-task internal diagnostics.

\subsection{ARC-TGI Generator Pipeline}
An ARC-TGI generator is a Python module implementing a transformation family. The \texttt{create\_input} stage samples grids with variation in color, position, size, and object configuration. The \texttt{transform\_input} stage deterministically applies the latent rule. The \texttt{create\_grids} stage assembles demonstrations and a test pair while rejecting samples that violate task constraints or leave the transformation ambiguous. Task-level variables remain fixed within an episode but vary across episodes; grid-level variables can vary among demonstrations from the same episode. Repeated calls therefore yield distinct instances with a shared transformation.

This structure enables resampling and controlled interventions that are unavailable in a fixed corpus. It also introduces a validity requirement: a generator must faithfully encode the intended ARC rule without permitting unintended shortcuts. Generator quality and the relationship between generated families and original ARC tasks are therefore part of the scope of the results.

The generator is consequently both the unit of data construction and a unit of
analysis. Episode-level exact match measures success on a sampled realization;
generator-level aggregation measures how broadly that success extends across
transformation families. We report both because a high aggregate score can be
driven by repeated success on a limited set of generators.

\subsection{Curriculum Construction and Splits}
The curriculum builder filters tasks according to experiment-specific criteria including token length, grid size, demonstration size, sample size, unique generators, and number of test examples per sample. Each record contains demonstration grids, a test grid, a task identifier, a generator identifier, and, where required by the prompt condition, a natural-language grid description and transformation trace. Training and evaluation records use disjoint task identifiers. Seen-family evaluation resamples from generators used in training; when used, novel-family evaluation relies on generator-disjoint splits. Spatial-transfer experiments additionally constrain the grid dimensions on each side of the split.

Seen-family resampling draws new episodes from a training generator under its
original sampling regime, whereas a within-family shift retains the generator
but changes a controlled parameter such as grid scale. The original ARC-AGI
evaluation set is treated separately as cross-benchmark transfer because it
changes both generator source and task difficulty; it is not a pure intervention
on novelty alone.

These splits support progressively stronger tests. Seen-family resampling asks
whether performance extends to new realizations of a trained transformation.
The spatial intervention asks whether it survives a controlled change in grid
scale. Cross-benchmark evaluation asks whether the selected checkpoint remains
useful on original ARC-AGI tasks, but does not isolate the source of any decline.

\subsection{Model-Selection Considerations}
The experimental volume required models that could be repeatedly fine-tuned on available H100 and H200 hardware. Selection also required instruction following, structured generation, and sufficient context for multiple grid demonstrations. The final representatives involve trade-offs rather than a perfectly matched comparison. Qwen3-1.7B provides a dense decoder-only baseline with a 32,768-token context. Qwen1.5-MoE-A2.7B-Chat provides sparse expert routing with approximately 14.3B total and 2.7B active parameters. T5Gemma-ml-ml-ul2-it provides an approximately 2B encoder--decoder representative with an 8K-token context. The MoE model therefore matches the intended active-compute scale more closely than the total-capacity scale, and the ED context is shorter than the initial 10K target. These deviations are reported explicitly rather than treated as controlled architectural differences.

\subsection{Training and Inference Infrastructure}
\label{app:reproducibility}
Training uses supervised fine-tuning with full-parameter or LoRA updates. Ray coordinates distributed execution on a SLURM cluster. NCCL timeout and error-handling variables are set to prevent silent distributed hangs. For MoE, target-module detection distinguishes expert feed-forward targets, which may require dynamic unused-parameter detection, from attention-only targets compatible with a static-graph path. The reported MoE LoRA runs target attention projections.

Inference uses Ray actors pinned to individual GPUs. vLLM is used for D and MoE when supported; ED uses the Hugging Face generation pipeline. Transduction answers are extracted from structured output tags. Induction answers are extracted as Python, executed in a sandbox, and evaluated by exact grid match. The reproducibility record includes model revisions, tokenizer revisions, random seeds, package versions, launch commands, dataset manifests, and checkpoint-selection rules.

Backend differences are recorded because they affect execution and resource
accounting, but no architectural conclusion is based on throughput or memory
alone. Behavioral comparisons use the same exact-match criterion after output
extraction or code execution, and compute measures are reported as properties of
the realized configurations rather than hardware-independent efficiency claims.

\section{RQ1: Optimization and Adaptation}
\label{app:optimization}
This section expands the operating-point analysis used before all acquisition
and transfer comparisons. It proceeds from the full hyperparameter response to
the compute and parameter-efficiency trade-offs of full fine-tuning and LoRA.
The purpose is not to declare a universally best optimizer setting, but to show
why each family requires its own empirically supported operating region before
the subsequent interventions can be discussed.

\subsection{Hyperparameter Search}
The initial sweep varies learning rate, warm-up ratio, weight decay, and epoch. Later stages fix the empirically dominant learning-rate region and refine the remaining factors. The evaluated grids are:

\begin{enumerate}
    \item D: learning rate $\{5\!\times\!10^{-6}, 10^{-5}, 2\!\times\!10^{-5}, 2\!\times\!10^{-4}\}$; warm-up ratio $\{0.03,0.05,0.1\}$; weight decay $\{0,0.01,0.05,0.1\}$; epochs $\{1,3,5,10,15,20,25,30,35\}$.
    \item MoE: learning rate $\{5\!\times\!10^{-6},10^{-5},2\!\times\!10^{-5},2\!\times\!10^{-4}\}$; warm-up ratio $\{0.03,0.05,0.1\}$; weight decay $\{0,0.01,0.05,0.1\}$; epochs $\{1,3,5,7,10,12\}$.
    \item ED: learning rate $\{10^{-5},5\!\times\!10^{-5},10^{-4},2\!\times\!10^{-4}\}$; warm-up ratio $\{0.03,0.05,0.1\}$; weight decay $\{0,0.01,10^{-4},10^{-5}\}$; epochs $\{1,3,5,10,15,20,25,30\}$.
\end{enumerate}

\begin{figure*}[t!]
  \centering
  \includegraphics[width=0.82\linewidth]{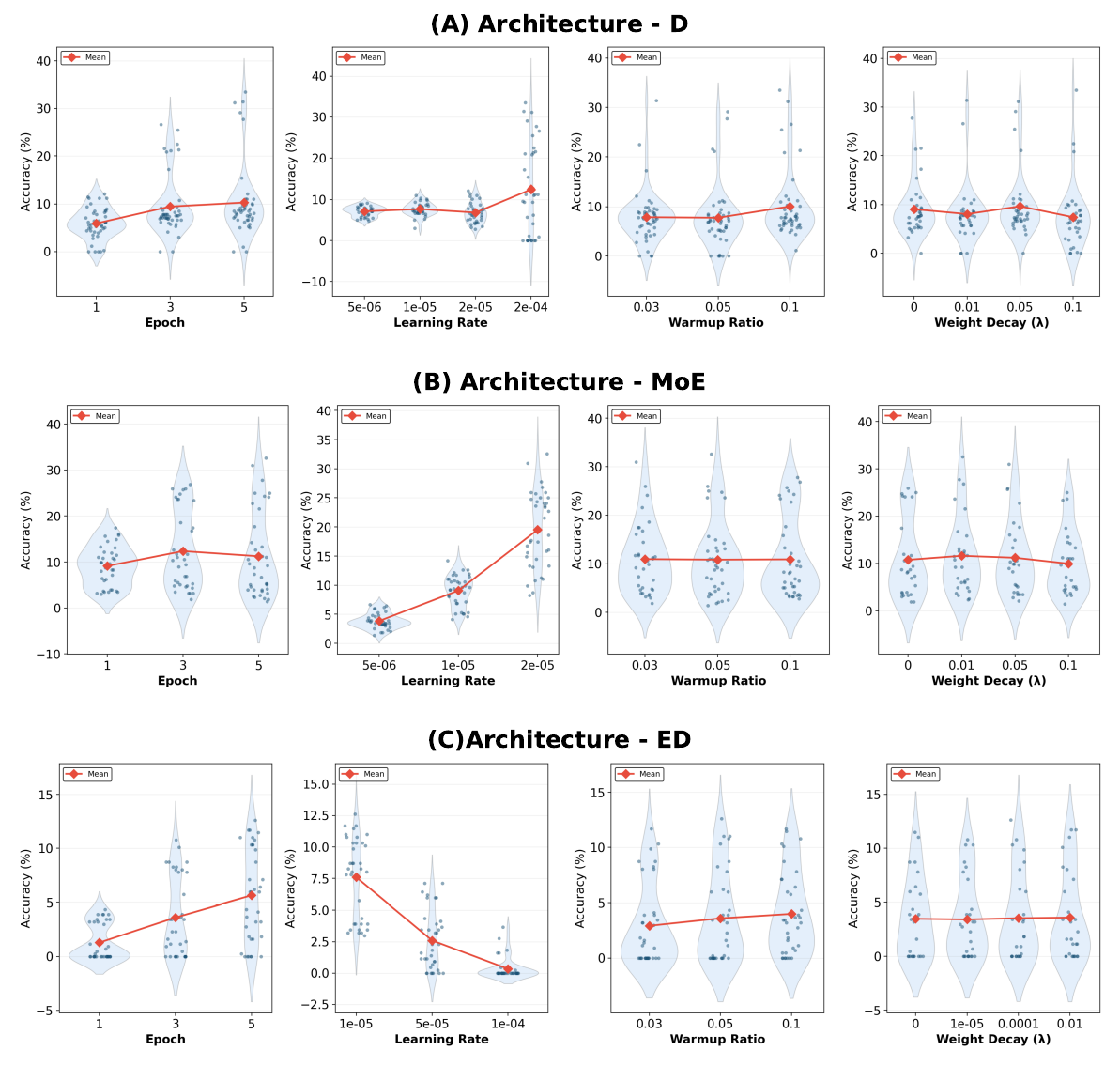}
  \caption{\textbf{RQ1: hyperparameter response distributions.} Exact-match
  accuracy across the tested epoch, learning-rate, warm-up, and weight-decay
  values for D (top), MoE (middle), and ED (bottom). Points are individual sweep
  configurations, violins summarize their marginal distributions, and red
  lines connect marginal means. Each panel marginalizes over the other
  hyperparameters and should therefore be read as a diagnostic response profile,
  not an isolated treatment effect.}
  \label{fig:rq1-marginals}
\end{figure*}

Figure \ref{fig:rq1-sensitivity} summarizes factor importance and the leading
configurations in the main text. Hyperparameter importance was quantified as the main-effect share of accuracy variance, $\eta^{2}_{A}=SS_{A}/SS_{T}$, where $SS_{A}=\sum_{a\in\mathcal{L}_{A}}n_{a}(\bar{y}_{a}-\bar{y})^{2}$ and $SS_{T}=\sum_{i=1}^{N}(y_{i}-\bar{y})^{2}$, expressed as a percentage \begin{equation} \mathrm{VE}_{A}\;=\;100\times\frac{SS_{A}}{SS_{T}}\;\in\;[0,100]\,\%, \end{equation} with $y_{i}$ the accuracy of run $i$, $\bar{y}$ the grand mean, and $\bar{y}_{a}$, $n_{a}$ the mean and number of runs at level $a$ of hyperparameter $A$.
For a fully crossed design, the importance scores of all hyperparameters sum to $\sum_{A}\mathrm{VE}_{A}\le 100\%$; any unexplained variance comes from interactions between hyperparameters. Figure \ref{fig:rq1-marginals} exposes the
underlying response distributions. D improves over the early epoch settings and
benefits from the largest tested learning rate, although most configurations
remain below the best operating region. MoE is dominated by learning rate:
accuracy rises sharply at $2\times10^{-5}$, while its warm-up and weight-decay
marginals remain comparatively flat. ED shows the opposite learning-rate
direction, with its strongest distribution at $10^{-5}$ and near-zero
performance at larger rates. Across families, the comparatively small shifts
associated with warm-up and weight decay support fixing these factors after the
initial search. Because each marginal averages over the other sampled factors,
these plots diagnose response shape rather than identify causal effects.

\subsection{Adaptation and Compute Accounting}
After locating an operating region for each family, we compare the selected
full-fine-tuning and LoRA runs in Table \ref{tab:app-adaptation-efficiency}.
Placing both methods in one table makes the relevant trade-off explicit: endpoint
accuracy must be read together with schedule length and realized resource use.
Adjusted GPU hours use the approximate conversion H200 $=2\times$ H100 for
accounting only; the conversion is not a hardware-equivalence claim.

\begin{table*}[t!]
\centering
\footnotesize
\setlength{\tabcolsep}{3.8pt}
\begin{tabular}{@{}llrrrrrrrr@{}}
\toprule
\textbf{Family} & \textbf{Update} & \textbf{Epoch} & \textbf{Acc.} & \textbf{Time} & \textbf{GPU} & \textbf{VRAM} & \textbf{Min/epoch} & \textbf{Adj. GPU h} & \textbf{Acc./h} \\
& & & (\%) & (sec) & & (GB) & & & \\
\midrule
D   & Full & 20 & 53.90 & 63303  & 4 H100 & 94  & 52.75  & 70.34 & 0.766 \\
    & LoRA & 25 & 54.36 & 80191  & 4 H100 & 94  & 53.46  & 89.10 & 0.611 \\
\addlinespace
MoE & Full & 5  & 32.57 & 111511 & 1 H200 & 141 & 371.70 & 61.95 & 0.526 \\
    & LoRA & 5  & 32.89 & 95580  & 1 H200 & 141 & 318.60 & 53.10 & 0.618 \\
\addlinespace
ED  & Full & 30 & 21.12 & 25365  & 4 H100 & 94  & 14.09  & 28.18 & 0.749 \\
    & LoRA & 10 & 10.81 & 16324  & 4 H100 & 94  & 27.21  & 18.14 & 0.595 \\
\bottomrule
\end{tabular}
\caption{\textbf{RQ1: unified adaptation and compute summary.} Selected
full fine-tuning and LoRA endpoints for each model family. Accuracy is exact
grid match; time and peak VRAM describe the realized run. Adjusted GPU hours use
the stated H200 $=2\times$ H100 accounting convention. Acc./h is reported
accuracy divided by adjusted GPU hours and is descriptive rather than a
hardware-independent efficiency measure.}
\label{tab:app-adaptation-efficiency}
\end{table*}

We evaluate $(r,\alpha)\in\{(16,32),(32,64),(64,128),(128,256)\}$ with dropout
0.05. Holding $\alpha/r=2$ keeps update scaling comparable while rank varies;
Figure \ref{fig:rq1-dynamics}(a) reports the resulting learning curves. The
selected LoRA run closely matches the full-fine-tuning endpoint for D and
slightly exceeds it for MoE, whereas ED retains roughly half of its
full-fine-tuning accuracy. Compute follows a different pattern: LoRA reduces
adjusted GPU hours for MoE and ED but increases them for D because its selected
schedule is longer. Thus, parameter-efficient adaptation is competitive for D
and MoE in accuracy and beneficial for MoE and ED in adjusted compute, but it is
not uniformly preferable on both criteria.

\section{RQ1: Extended Acquisition and Experience Analyses}
\label{app:experience}
This section moves from training time to training data: checkpoint trajectories
characterize stability, within-family depth tests repeated variation of a fixed
rule set, and breadth tests exposure to additional transformation families. The
three analyses use different units of progress---updates, samples per generator,
and number of generators---but share the same question: whether aggregate gains
represent stable and broadly distributed acquisition.

\subsection{Checkpoint-Level Learning Dynamics}
For D and ED, 20 checkpoints are sampled uniformly over 10 epochs; for MoE, 20 checkpoints are sampled over five epochs. Early stopping is disabled. Each checkpoint is evaluated on held-out seen-family data with 20 examples per generator.
For generator-level performance statistics, we calculate $\Delta$ as the difference between the final and initial accuracy,
\[
\Delta = A_{\mathrm{final}} - A_{\mathrm{initial}}.
\]
Generators with $\Delta > 0$ are categorized as improved, whereas those with $\Delta < 0$ are categorized as decreased. Zero-performing generators are those whose accuracy remains zero across all sampled checkpoints.

\begin{figure*}[t!]
  \centering
  \includegraphics[width=\linewidth]{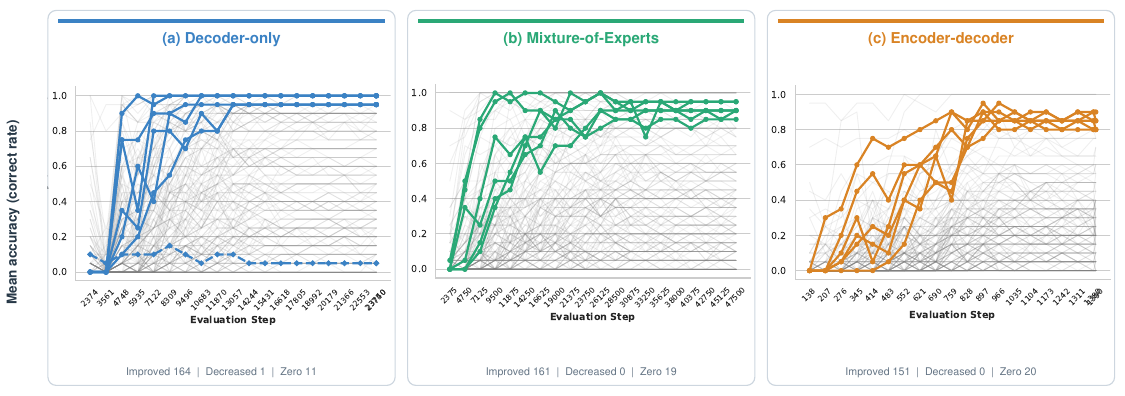}
  \caption{\textbf{RQ1: generator-level checkpoint dynamics.} Held-out
  exact-match trajectories for D (blue), MoE (green), and ED (orange) across 20
  checkpoints. The six largest improvements are marked with solid circles, and
  three representative declines or zero-change cases with dashed diamonds;
  remaining generator trajectories are gray. Counts below each panel summarize
  improved, decreased, and zero-performing families between the sampled
  endpoints.}
  \label{fig:rq1-checkpoints}
\end{figure*}

Figure \ref{fig:rq1-checkpoints} resolves the aggregate curves in Figure
\ref{fig:rq1-dynamics}(b) into generator-level trajectories. D exhibits rapid
early acquisition but substantial heterogeneity across families; 164 families
improve, one decreases, and 11 remain at zero performance. MoE similarly shows
rapid acquisition, with 161 improving families, no decreasing cases, and 19
zero-performing families, whereas ED improves more gradually and retains the
largest zero-performing set (20 families), despite 151 families improving. Abrupt transitions from failure to high accuracy further show that the aggregate curves do not represent
smooth progress on a typical generator. Since uniformly sampled checkpoints do
not necessarily include the trainer-selected optimum, this analysis
characterizes acquisition stability rather than replacing endpoint selection.

\subsection{Experience Depth and Breadth}
\begin{figure*}[t!]
  \centering
  \includegraphics[width=\linewidth]{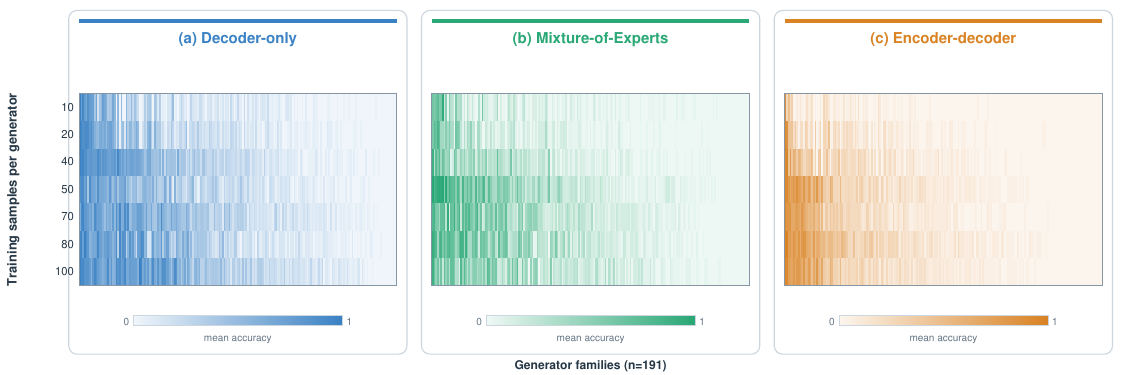}
  \caption{\textbf{RQ1: experience depth.} Generator-level exact-match
  accuracy across 191 fixed transformation families as training examples per
  family increase from 10 to 100. Panels correspond to D (blue), MoE (green),
  and ED (orange). Columns are generators, rows are sample counts, and intensity
  encodes accuracy on a shared 0--1 scale. Persistent vertical structure shows
  that generator identity remains influential as within-family experience
  increases.}
  \label{fig:rq1-sample-scaling}
\end{figure*}

Within-family depth is varied over
$n\in\{10,20,40,50,70,80,100\}$ training samples per generator. The same 200
generator identities are used across training scales, and evaluation uses a
fixed, disjoint support of 30 examples for each of 191 retained generators. This design isolates repeated experience with the same
transformation families. Figure \ref{fig:rq1-sample-scaling} shows that D and
MoE contain more high-accuracy families than ED, but all three profiles retain
strong vertical structure: generators that are easy or difficult at small
sample counts often remain so. Non-monotonic cells additionally show that more
examples can coincide with either acquisition or forgetting. Depth therefore
changes individual families without producing a uniform expansion of the
learnable set.

\begin{figure*}[t!]
  \centering
   \includegraphics[width=\linewidth]{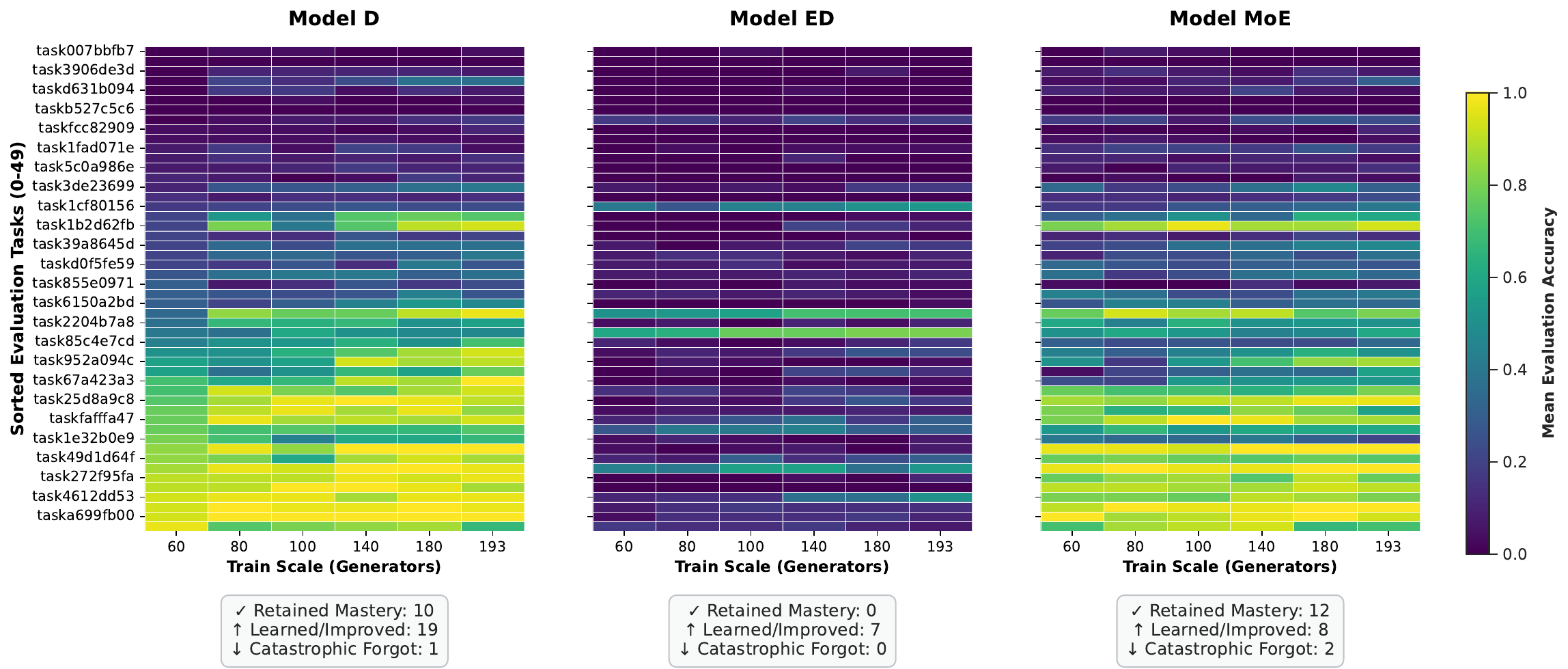}
  \caption{\textbf{RQ1: experience breadth.} Generator-level exact-match
  accuracy on 50 fixed evaluation families as the training curriculum expands
  from 60 to 193 generators. From left to right, panels show D, MoE, and ED.
  Columns are curriculum sizes and rows are shared evaluation
  generators, sorted independently by each model's 60-generator baseline.
  Color encodes accuracy on a common 0--1 scale; annotations summarize retained,
  improved, and markedly decreased endpoint cases.}
  \label{fig:rq1-generator-scaling}
\end{figure*}

Training breadth is varied over generator sets of size
$\{60,80,100,140,180,193\}$ with 50 training examples per generator and 20
evaluation examples per generator. The evaluation support is fixed to 50
generators shared across all scales, so changes cannot be attributed to a
different test composition. Figure \ref{fig:rq1-generator-scaling} shows that
the average improvement in Table \ref{tab:app-generator-scaling} is not uniform:
D improves on 19 shared families while one previously strong family declines;
MoE improves on 8, retains 12, and declines on two; and ED improves on seven
but otherwise remains close to its low baseline. The intermediate columns also
contain reversals, indicating that breadth improves the aggregate endpoint
without producing monotonic progress for every transformation family.

\begin{table}[t!]
\centering
\small
\begin{tabular}{lccc}
\toprule
\textbf{Generators} & \textbf{D} & \textbf{MoE} & \textbf{ED} \\
\midrule
60  & 39.3 & 39.1 & 7.5 \\
80  & 44.8 & 39.2 & 8.7 \\
100 & 43.8 & 41.5 & 9.7 \\
140 & 49.3 & 43.2 & 12.7 \\
180 & 51.0 & 45.1 & 12.2 \\
193 & 51.6 & 45.0 & 13.0 \\
\bottomrule
\end{tabular}
\caption{\textbf{RQ1: aggregate experience-breadth results.} Exact-match
accuracy (\%) on the same 50 evaluation families as the training curriculum
expands from 60 to 193 generators. These averages summarize the same conditions
shown at generator level in Figure \ref{fig:rq1-generator-scaling}.}
\label{tab:app-generator-scaling}
\end{table}

Depth and breadth therefore produce the same high-level result through different
interventions. Additional within-family samples can change individual
generators without removing persistent easy and difficult groups, while broader
curricula improve the shared evaluation average through gains on only part of
the support. Together with the checkpoint trajectories, these results explain
why RQ1 is stated in terms of both acquisition efficiency and stability.

\section{RQ2: Extended Transfer Analyses}
\label{app:generalization}
This section separates two transfer tests from one coverage analysis.
Cross-benchmark evaluation measures performance on original ARC-AGI tasks but
changes several aspects of the distribution simultaneously. Spatial-scale
transfer retains generator identity and changes the grid regime. The final
analysis returns to held-out instances of seen families to determine how broadly
the reference performance is distributed; it is not itself an additional
transfer condition. Table \ref{tab:app-transfer-details} presents the complete
numerical results for the two transfer tests.

\subsection{Cross-Benchmark Transfer over Training}
Table \ref{tab:app-transfer-details}(a) reports exact-match accuracy on the
original ARC-AGI evaluation set for every available checkpoint. The trajectories
fluctuate and often peak before the final seen-family optimum, so checkpoint
selection based only on held-out seen-family data can miss the strongest
cross-benchmark result. The reported maxima are identified after inspecting these
trajectories and are therefore descriptive rather than independently selected
test estimates.

D reaches its cross-benchmark maximum at epoch 25, MoE at epochs 1 and 5,
and ED at epochs 15 and 20. Later seen-family checkpoints do not preserve these
maxima: every family exhibits a decline followed by at most partial recovery.
The trajectory therefore reinforces the main-text conclusion that checkpoint
quality on seen generators is not a monotonic proxy for transfer. Since the
target benchmark also changes task difficulty and generator source, these values
remain evidence of cross-benchmark transfer rather than a clean novelty effect.

\subsection{Spatial-Scale Transfer}
The spatial intervention retains 123 generator families and changes only the
allowed grid-size regime. Each training curriculum contains 50 examples per
generator, and each evaluation curriculum contains 20 disjoint examples for each of 118 generators. Table \ref{tab:app-transfer-details}(b) presents the complete matched
and crossed matrix underlying the RQ2 spatial rows of Table
\ref{tab:main-summary}. Unlike the cross-benchmark evaluation, this comparison
retains the transformation families and changes a specified property of their
realization.

Matched-scale evaluation is strongest for every family with one exception. The small-to-large
penalty is 44.2 points for D, 40.4 for MoE, and 58.5 for ED; the corresponding
large-to-small penalties are smaller at 13.6 and 6 for D and MoE, respectively. However, ED demonstrates opposite behavior, increasing large-to-small performance by 11.1 points. This
directional asymmetry shows that exposure to longer serialized grids transfers
better to shorter grids than the reverse for the D and MoE models. Context length and representation are
plausible contributors, particularly for ED, but the model-family comparison
does not isolate either mechanism causally.
\subsection{Seen-Family Coverage and Heterogeneity}
For each of 191 generators, we sample up to 30 evaluation instances disjoint
from training and aggregate exact-match accuracy by family. Figure
\ref{fig:rq2-heterogeneity} shows a steep profile: D exceeds 80\% on
approximately 50 generators, MoE on approximately 25, and ED on approximately
10. D combines the highest mean with the widest dispersion, whereas ED's lower
dispersion partly reflects widespread failure. Thus, the aggregate reference
accuracy does not imply broad coverage of seen transformations. We report this
as \emph{family-level heterogeneity}: cross-generator dispersion measures neither
consistency across a rule's parameter settings nor transfer to an unseen family.

\begin{table*}[t!]
\centering
\small
\begin{minipage}[t]{0.42\linewidth}
\centering
\textbf{(a) Cross-benchmark checkpoints}\par\smallskip
\begin{tabular}{lrrr}
\toprule
\textbf{Epoch} & \textbf{D} & \textbf{MoE} & \textbf{ED} \\
\midrule
0  & 0.5  & 0.25 & 0.25 \\
1  & 2.0   & 2.75 & 1.0 \\
3  & 3.0   & 1.25 & 1.75 \\
5  & 0.5   & 2.75 & 1.25 \\
7  & --    & 1.50 & -- \\
10 & 0.25 & --   & 1.50 \\
15 & 2.0   & --   & 2.0 \\
20 & 1.0   & --   & 2.0 \\
25 & 3.50  & --   & 0.25 \\
30 & 2.50  & --   & 1.25 \\
\bottomrule
\end{tabular}
\end{minipage}
\hfill
\begin{minipage}[t]{0.54\linewidth}
\centering
\textbf{(b) Spatial-scale transfer matrix}\par\smallskip
\begin{tabular}{llrr}
\toprule
\textbf{Family} & \textbf{Training grids} & \textbf{Small eval.} & \textbf{Large eval.} \\
\midrule
D   & Small ($\leq15$) & 68.5 & 24.3 \\
    & Large ($\geq16$)    & 51.4 & 65 \\
\addlinespace
MoE & Small ($\leq15$) & 53.1 & 12.7 \\
    & Large ($\geq16$)    & 36.4 & 42.4 \\
\addlinespace
ED  & Small ($\leq15$) & 64.4 & 5.9 \\
    & Large ($\geq16$)    & 45 & 33.9 \\
\bottomrule
\end{tabular}
\end{minipage}
\caption{\textbf{RQ2: unified transfer results.} Exact-match accuracy (\%).
Left: original ARC-AGI evaluation at each available checkpoint; dashes denote
unmeasured checkpoints. Right: matched and crossed grid-scale evaluation with
generator identity shared across conditions. The left panel is a
cross-benchmark comparison, whereas the right panel isolates a spatial shift
within the filtered ARC-TGI families.}
\label{tab:app-transfer-details}
\end{table*}

\section{RQ3: Task Formulation and In-Context Experience}
\label{app:formulation}
This section separates two sources of expressed capability: the requested
output representation and the amount of task evidence supplied in the prompt.
The few-shot condition supplies demonstration pairs and asks for either an output grid or Python code. The reasoning-description condition additionally supplies natural-language descriptions and transformation traces. Because the best configuration from the original transduction sweep did not transfer reliably across formulations, the formulation study performs a smaller search over warm-up ratio and weight decay, with epochs fixed at 10 for D and ED and five for MoE. Full-parameter fine-tuning is used to limit the remaining search space.

For the demonstration-count intervention, task and generator identifiers are held constant across $k\in\{1,2,3,4\}$. The evaluation contains approximately 5,800 examples from 186 generators. Figure \ref{fig:rq3-shots} provides the generator-level pairwise comparisons underlying the aggregate values in Table \ref{tab:rq3-formulation}.

\begin{table}[h!]
\centering
\small
\setlength{\tabcolsep}{4.5pt}
\begin{tabular}{@{}llrrr@{}}
\toprule
\textbf{Intervention} & \textbf{Condition} & \textbf{D} & \textbf{MoE} & \textbf{ED} \\
\midrule
\multirow{4}{*}{Formulation}
 & Transduction: few-shot  & 44.3 & 33.7 & 0.0 \\
 & Transduction: reasoning & 42.7 & 32.6 & 15.6 \\
 & Induction: few-shot     & 52.3 & 59.0 & 16.1 \\
 & Induction: reasoning    & 42.4 & 59.2 & 12.6 \\
\midrule
\multirow{4}{*}{Context}
 & One shot   & 36.7 & 20.1 & 43.6 \\
 & Two shots  & 45.8 & 26.5 & 37.9 \\
 & Three shots & 49.5 & 28.3 & 27 \\
 & Four shots & 51.9 & 30.6 & 23.4 \\
\bottomrule
\end{tabular}
\caption{\textbf{RQ3: formulation and in-context experience.} Exact-match
accuracy (\%) under direct grid transduction and executable-rule induction with
few-shot or reasoning-description prompts, followed by accuracy on fixed task
and generator support as demonstrations increase from one to four.}
\label{tab:rq3-formulation}
\end{table}

The formulation results separate output representation from added natural
language. Induction improves few-shot performance across all three model representatives,
raising D from 44.3\% to 52.3\%, MoE from 33.7\% to 59.0\%, and rescuing ED
from 0.0\% to 16.1\%. Reasoning descriptions have mixed effects: under
transduction they slightly reduce D and MoE but raise ED to 15.6\%, whereas
under induction they reduce D and ED while leaving MoE essentially unchanged.
Thus, reasoning descriptions do not act as a uniform source of supervision
across model representatives or output formulations.
The demonstration-count intervention yields a second interaction: D and MoE improve
monotonically from one to four examples, while ED declines from 43.6\% to
23.4\%. Figure \ref{fig:rq3-shots} shows that this reversal is distributed
across transformation families rather than produced by a single outlier.

The two interventions should not be collapsed into a single prompt effect.
Changing from transduction to induction changes the target representation and
introduces an executable-language prior, whereas increasing shot count changes
the amount and length of task evidence. Their family-specific interactions show
that the evaluation interface can alter both the measured endpoint and the set
of transformations on which a model succeeds. They do not establish that one
formulation is intrinsically easier across models or task distributions.

\section{RQ4: Exploratory Internal-Signature Case Studies}
\label{app:mechanistic}
This section provides the definitions and scope of the two diagnostics applied
to selected cases and reported together in Figure \ref{fig:rq4-diagnostics}. They provide
hypotheses about where model-family profiles diverge, but neither is used as
primary evidence for the behavioral conclusions.

\subsection{Attention Entropy}
For selected small- and large-grid tasks, we compute Shannon
entropy from the attention distribution over key positions, separately for every
layer, head, and retained query position. The retained positions and
normalization domains differ by attention type, as illustrated in Figure
\ref{fig:rq4-diagnostics}. Entropy is reported in nats as raw per-layer
values; no smoothing is applied. The two variants therefore have different maximum-entropy ceilings (the logarithm of the prompt length for
self-attention in D and MoE versus the logarithm of the test-region length for cross-attention in ED). Lower values indicate a
more concentrated distribution, but do not identify the attended content or
imply better task performance.

The selected D and MoE cases show decreasing entropy with depth, with only modest differences between small- and large-grid training. By contrast, the selected ED cases show the clearest separation between training regimes from the middle layers onward. These profiles are consistent with different attention-concentration behavior, but the small task sample and family-level confounds preclude general architectural conclusions.

\subsection{Attention-Edge Knockout} \label{app:attention-knockout}
We partition the prompt into system instructions, demonstrations, and the
test-grid description. For each prompt region and five-layer window, we mask
incoming attention edges from that region to the final query
position (self-attention) or to all decoder query positions (ED
cross-attention), without removing prompt tokens.
Let $\bar{\ell}_{0}$ denote the mean per-token target-answer log
probability of the unmodified pass, averaged over the $A$ answer tokens under
teacher forcing, and let $\bar{\ell}_{m}$ denote the same quantity after
masking. The reported effect is the relative change: 

\begin{equation} \label{app:eq:logpchange}
\Delta_{\%} \;=\; \frac{\bar{\ell}_{m} - \bar{\ell}_{0}}
                       {\lvert \bar{\ell}_{0} \rvert} \times 100 .
\end{equation}

Because the denominator is positive, $\Delta_{\%}$ preserves the
sign of the underlying log-probability change. Negative values therefore mean
that removing the region lowers target probability and the region was
supportive in that diagnostic pass; positive values mean that masking removes a
competing signal. The bottom half of Figure \ref{fig:rq4-diagnostics} reports
these effects on the selected tasks using per-family baseline-normalized scales.
The normalization in Equation \ref{app:eq:logpchange} rescales each family by
$\lvert \bar{\ell}_{0} \rvert$, which differs across models: an identical
absolute change in log probability yields a larger $\Delta_{\%}$ for a model
whose unmodified prediction is more confident. Reported magnitudes are
therefore interpretable within a panel but not across panels.

The selected decoder-only case exhibits a layer-dependent sign change in the contribution of demonstrations. The selected MoE case shows a strong dependence on system message over an intermediate layer range. Encoder self-attention interventions are noisier, while decoder cross-attention provides a more interpretable signal. These observations motivate larger causal studies but are not used as primary evidence in the main paper.

The diagnostics are complementary rather than interchangeable. Entropy measures
how dispersed attention is but does not identify which prompt region supports
the answer. Edge knockout tests the contribution of a specified region, but its
effect depends on the selected prompt and layer window. Their joint value is to
separate concentration from context dependence on diagnostic cases; establishing
a general mechanism would require a larger task sample, repeated interventions,
and matched representatives within each model family.

\end{document}

%% file: math_commands.tex
\usepackage{amsmath,amsfonts,bm}

\def\eqref#1{equation~\ref{#1}}

\def\1{\bm{1}}

\DeclareMathAlphabet{\mathsfit}{\encodingdefault}{\sfdefault}{m}{sl}
\SetMathAlphabet{\mathsfit}{bold}{\encodingdefault}{\sfdefault}{bx}{n}

